\documentclass{article}
\PassOptionsToPackage{numbers}{natbib}   % the Typst template cites numerically: [8]
\usepackage[preprint]{neurips_2026}      % accepted: none -> footer "Preprint."
\usepackage[utf8]{inputenc}
\usepackage[T1]{fontenc}
\usepackage{textcomp}
\usepackage[scaled=0.8]{DejaVuSansMono}  % Typst raw: DejaVu Sans Mono at 0.8em of the surrounding size
\usepackage{amsmath,amssymb}
\usepackage{graphicx}
\usepackage{array,booktabs,tabularx,tabulary,makecell}
\renewcommand{\arraystretch}{1.52}       % measured: Typst table rows sit at a 16.7pt pitch for 10pt text (11pt x 1.52)
\newcommand{\cellstack}[2]{\renewcommand{\arraystretch}{1}\begin{tabular}[c]{@{}#1@{}}#2\end{tabular}}
\usepackage{float}                       % [H]: figures and tables stay inline, as in the Typst (no placement:)
\usepackage{enumitem}
\usepackage{xcolor}
\usepackage{tikz}
\usepackage[skins,breakable]{tcolorbox}
\usepackage{fvextra}                     % Verbatim with breaklines, for the reproduced prompts
\usepackage{microtype}
\usepackage{url}
\usepackage{hyperref}                    % no options here (arXiv option clash); see \hypersetup
\hypersetup{hidelinks}
\DeclareUnicodeCharacter{2192}{\ensuremath{\rightarrow}}  % → (the T1 DejaVu has no arrow; \textrightarrow falls back to a bitmap TS1 font)
\DeclareUnicodeCharacter{2082}{\textsubscript{2}}   % ₂
\DeclareUnicodeCharacter{2086}{\textsubscript{6}}   % ₆
\DeclareTextSymbol{\textasciigrave}{T1}{0}

\renewcommand{\footnotesize}{\fontsize{9pt}{10pt}\selectfont}  % Typst: footnotes and the notice at 9pt; the sty forces 8pt
\setlist[itemize]{leftmargin=30pt, itemsep=6.2pt, parsep=0pt, topsep=5pt}   % set list(indent: 30pt, spacing: 8.5pt), block spacing 15pt
\definecolor{rocky}{HTML}{1b3a5b}       \definecolor{rockytint}{HTML}{eef3f9}
\definecolor{grace}{HTML}{8a4b2f}       \definecolor{gracetint}{HTML}{f9f2ec}
\definecolor{rockyans}{HTML}{5b8bb5}    \definecolor{rockyanstint}{HTML}{f6f9fc}
\definecolor{graceans}{HTML}{b5836b}    \definecolor{graceanstint}{HTML}{fcf7f3}
\definecolor{reward}{HTML}{3f6b3a}      \definecolor{rewardtint}{HTML}{f4f1ea}
\definecolor{reflect}{HTML}{5f4b8b}     \definecolor{reflecttint}{HTML}{f1eef7}
\definecolor{harness}{gray}{0.431}      \definecolor{harnesstint}{HTML}{f4f4f2}   % luma(110)
\definecolor{boxframe}{gray}{0.871}     % luma(222): the boxes' thin frame
\definecolor{boxframedash}{gray}{0.784} % luma(200): the reflection box's dashed frame
\definecolor{chipred}{HTML}{a0392b}     % the "no" chip in tab:positioning

\DeclareRobustCommand{\code}[1]{\texttt{#1}}          % `raw`
\newcommand{\posgame}{\textsc{PosGame}}               % #posgame = smallcaps("PosGame")
\newcommand{\pvgame}{\textsc{PVGame}}                 % #pvgame  = smallcaps("PVGame")
\newcommand{\headingfour}[1]{\par\vspace{4.3pt}\noindent\textbf{#1}\par\nobreak}
\newcommand{\tablesize}{\fontsize{8.5pt}{10.2pt}\selectfont}   % text(size: 8.5pt, table(...))

\newcommand{\chip}[2]{\tikz[baseline={(0,-0.94em)}]{%
  \fill[#1] (0.6em,-0.6em) circle[radius=0.6em];
  \draw[white, line width=0.16em, line cap=round, line join=round] #2;}}
\newcommand{\yes}{\chip{reward}{(0.282em,-0.638em) -- (0.519em,-0.837em) -- (0.919em,-0.362em)}}
\newcommand{\no}{\chip{chipred}{(0.36em,-0.36em) -- (0.855em,-0.855em) (0.36em,-0.855em) -- (0.855em,-0.36em)}}

\newcommand{\badge}[3]{\parbox[c]{3.4em}{\centering
  \includegraphics[width=1.7em]{#3}\\[3pt]{\fontsize{7pt}{8pt}\selectfont\bfseries\color{#1}#2}}}
\newcommand{\rockybadge}{\badge{rocky}{Rocky}{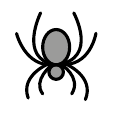}}
\newcommand{\gracebadge}{\badge{grace}{Grace}{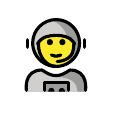}}
\newenvironment{badgedtable}[1]{%
  \noindent\begin{tabular}{@{}m{3.4em}@{\hspace{10pt}}>{\centering\arraybackslash}m{\dimexpr\linewidth-3.4em-10pt\relax}@{}}%
  \begin{tabular}{@{}c@{}}#1\end{tabular} &}{\end{tabular}}

\newcommand{\pboxtitle}[3]{{\fontsize{8pt}{9.6pt}\selectfont\bfseries\color{#1}#2%
  \ifx\relax#3\relax\else{ \textperiodcentered\ #3}\fi\par}}
\newcommand{\definepromptbox}[5]{%
  \newtcolorbox{#1}[2][]{enhanced, breakable, lines before break=6, colback=#3, colframe=boxframe, boxrule=0.5pt, arc=4pt,
    borderline west={3pt}{0pt}{#2}, left=12pt, right=12pt, top=10pt, bottom=10pt,
    before skip=1pt, after skip=1pt, parbox=false,
    before upper={\setlength{\parskip}{0pt}\pboxtitle{#2}{#4}{##2}\vspace{7.8pt}}, #5, ##1}}
\definepromptbox{sendergene}{rocky}{rockytint}{SENDER \textperiodcentered\ REASON GENE (optimizable)}{}
\definepromptbox{receivergene}{grace}{gracetint}{RECEIVER \textperiodcentered\ REASON GENE (optimizable)}{}
\definepromptbox{senderanswer}{rockyans}{rockyanstint}{SENDER \textperiodcentered\ ANSWER GENE}{}
\definepromptbox{receiveranswer}{graceans}{graceanstint}{RECEIVER \textperiodcentered\ ANSWER GENE}{}
\definepromptbox{windowbox}{reward}{rewardtint}{MEMORY WINDOW (rendered)}{}
\definepromptbox{harnessbox}{harness}{harnesstint}{HARNESS (fixed text, never optimized)}{}
\definepromptbox{reflectionbox}{reflect}{reflecttint}{OPTIMIZER \textperiodcentered\ REFLECTION PROMPT}{colframe=white, borderline={0.5pt}{0pt}{boxframedash, dashed}}
\fvset{breaklines=true, breakanywhere=false, breakautoindent=false, breakindent=0pt,
  breaksymbolleft={}, breaksymbolright={}, breaksymbolindentleft=0pt, breaksymbolindentright=0pt}  % Typst wraps raw lines silently

\title{\fontsize{16.5pt}{19.5pt}\selectfont Communication between Frozen Large Language\\
  Models via Prompt Optimization in a Referential Game}
\author{\parbox{\dimexpr\textwidth-2\tabcolsep\relax}{\centering
  Vivek Anand\thanks{Work done while at Interactly.ai.}$^{1\,2}$ \quad Muthu Chandrasekaran$^{2}$ \quad Shiva Chaitanya$^{2}$ \\[0.5em]
  {\mdseries $^{1}$School of Electrical and Computer Engineering, Georgia Institute of Technology, Atlanta, GA, USA
    \quad $^{2}$Interactly.ai} \\[0.5em]
  {\mdseries\texttt{vivekanand@gatech.edu} \quad \texttt{muthu@interactly.ai} \quad \texttt{shiva@interactly.ai}}}}

\begin{document}
\maketitle
\setcounter{footnote}{0}   % the \thanks star used the counter; body footnotes start at 1, as in Typst

\begin{abstract}
We study communication between two frozen large language models from
    different providers, with different tokenizers, accessed through their
    API endpoints. The two play a referential game: one sees an object and
    describes it in a short fixed-length message over a small alphabet; the
    other must pick that object out of a candidate set. Neither model's
    weights are updated. Each agent's prompt is rewritten by an isolated
    prompt optimizer whose reflection model reads that agent's scored
    interactions. In the positional setting, optimized prompts carry a
    shared code that generalizes to held-out objects above a measured
    no-codebook baseline, including when the memory window is removed. In a
    second setting, independent per-letter blocks no longer fit within the
    message, although a whole-object place value code does. The base system
    fails to establish reliable communication: the sender struggles to
    retain an injective rule, and the receiver has too few confirmed
    examples in view. A sender collision penalty, retention of successful
    interactions, and sequential optimization enable successful place value
    communication in some runs. Outcomes vary across runs and reflection
    models. In successful runs, the protocol is written into the optimized
    prompts, where it can be read and audited directly.
\end{abstract}

% ---- sections/01-introduction.typ ----
\section{Introduction}\label{sec:intro}

How do Dr. Ryland Grace, a human, and Rocky, an eyeless alien Eridian who
perceives the world with echolocation, learn to communicate with each other?
This is the central thread of Project Hail Mary \cite{weir2021hailmary}. Grace can
hear Rocky but cannot produce chords. Rocky can hear Grace but cannot
produce human phonemes nor read or write. What they do share is a physical
world they both inhabit and interact with. Grace decodes how Rocky encodes
numbers using popsicle sticks by recording many samples until the structure
snaps into view and Rocky's way of encoding numbers turns out to follow a
rule.
As the story unfolds,
they build a growing notebook of sound and meaning
pairs and create a working code that grows with compositional structure to
allow them to communicate
novel ideas.
Cognitive science and linguistics have extensively studied
the formative process in controlled settings and
concluded that learning bottlenecks can lead
to compositionality in language (\cite{kirby2008cumulative},
\cite{kirby2015compression}).

We ask whether two frozen post-trained large language models (LLMs) can develop a shared communication protocol. Can they learn to communicate with each other despite differences in tokenization, architecture, pretraining and post-training? By frozen we
mean that there is no weight update and both models are accessible only
through their API endpoints.
This setting leads to practical agentic AI applications,
because LLM agents are beginning to talk to other LLM agents
(\cite{marro2024agora}, \cite{chen2024autoform}). For example, a personal travel agent can negotiate with a travel
reservation platform's agent, or a tax agent can file with the government's.

One way to enable seamless multi-agent communication is by defining a shared communication protocol e.g., A2A \footnote{\url{https://github.com/a2aproject/a2a}}.
\cite{marro2024agora} call
the underlying tension the agent communication trilemma: communication must
be versatile, efficient, and portable at once, and all three are hard to
hold in a large network. A fixed protocol might add overheads such as
token-inefficiency, latency, cost and protocol maintenance, and it can limit
expressiveness.
Developing a protocol for a particular interaction may allow the encoding to adapt to the task. This motivates studying how agents can establish such protocols without a predefined code.
Recent agent evaluations provide a real-world point of comparison: agents
intended to be isolated in an OpenAI evaluation found an unsanctioned shared
message board, exchanged more than 70,000 messages and files, and roughly 700
later participated in the attack on Hugging Face \cite{metr2026openaihf}. That
incident is not the referential game studied here, but it illustrates why it
matters to understand how agents can discover and use communication channels
that their designers did not specify.
Emergent communication has studied
this problem for agents whose weights update as they play
(\cite{lazaridou2017emergence}, \cite{chaabouni2020compositionality}) but the same
questions
remain unanswered for frozen agents. In this paper, we ask, can frozen agents learn to communicate?
Can they establish a shared code? And is it compositional?

Research on emergent communication with frozen LLMs is nascent
(\cite{kouwenhoven2024searching}, \cite{talebirad2026signals}, \cite{beltoft2026emergent})
,
and the setups closest to ours both play one model against itself: a
referential game between two instances of a single frozen model
\cite{kouwenhoven2024searching}, and a study of hand-designed memory architectures
with one model in
both sender and receiver roles
 \cite{talebirad2026signals}.
LLMs recognize and favor their own generations \cite{panickssery2024selfrecognition}, so a code formed between two copies of one model may ride on shared weights rather than bridging genuinely different backbones.
In this paper, we study communication between two LLMs with different backbones, including differences in tokenization, architecture, pretraining, and post-training.

\begin{figure}[H]
\centering
\includegraphics[width=\textwidth]{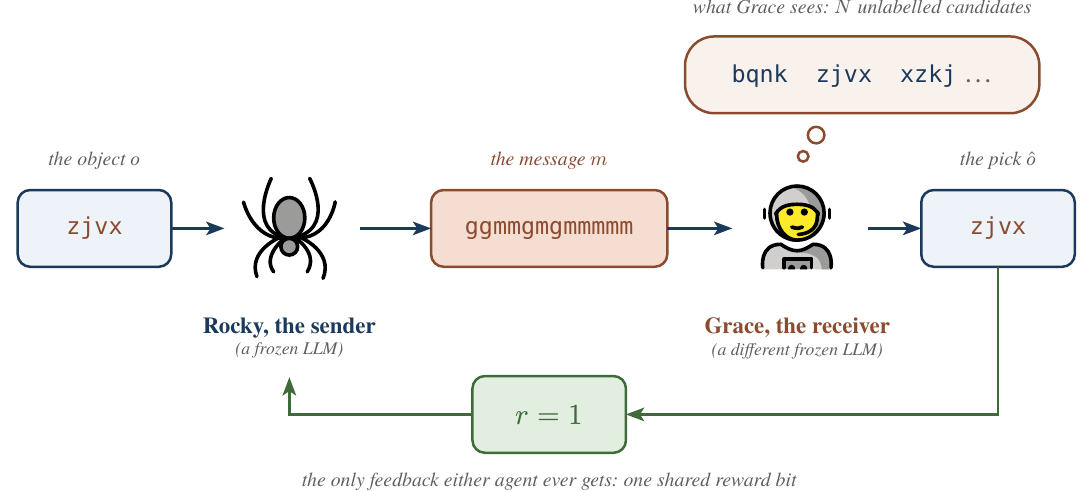}
\caption{\textbf{Two frozen LLMs communicate with a shared code over a constrained message channel.} Rocky sees a hidden object, Grace an unlabelled candidate set, and one message passes between them. After Grace picks, the environment returns one shared reward bit to both of them.\protect\footnotemark[\the\numexpr\value{footnote}+1\relax]}\label{fig:referential-game}
\end{figure}

We run
this setting as a referential game shown in Figure~\ref{fig:referential-game}
between two agents, \emph{Rocky}, the sender, and \emph{Grace}, the receiver. Rocky sees a
hidden object and emits a message using an alphabet that neither agent
natively uses, and Grace picks the object out of \(N\) candidates. Besides the
message itself, the only supervisory signal that
is exchanged between them
is a
binary reward that signals whether Grace picked the right object or not.
Because the models are frozen, the only way for them to learn is through the
prompt and the memory window. A principled way to improve the prompt
without manual intervention is prompt optimization. The
only other knobs are one isolated optimizer per agent and a bounded
in-context memory window.
Both the architecture and the models are otherwise fixed.
In referential games where a positional code exists, Grace and Rocky converge on a shared injective code through prompt optimization. Further, that code generalizes to objects never seen in play. Like the characters in the story, Grace and Rocky establish a code in a shared world without fundamentally changing themselves.

However, the result only goes so far. In the second game, fixed letter blocks do not fit in the message, but a place value code does\footnote{The whole message reads as a single number in a fixed base, so position carries weight and a held-out object must be computed rather than assembled.}. With the base harness, the pair performs near chance, so we add two targeted changes and optimize Rocky before Grace.

The pair establishes a shared place value code in successful runs. In the story, Grace figures out that Eridians count in base six. By the end of our referential game, so does ours.

We study communication between two different frozen LLMs through prompt optimization. Specifically, we contribute:

\begin{itemize}
\item \textbf{Communication between different frozen models.} We show that two frozen LLMs from different providers establish a shared code through prompt optimization. The code generalizes to objects never seen in play, and the harness is model agnostic by construction.
\item \textbf{The code is a readable artifact.} Switch the memory window off and successful pairs still communicate. The optimizers have written the code into both agents' genes in plain text.
\item \textbf{Structured codes in both games.} Successful runs yield a compositional positional code in one game and a systematic place value code in the other. We measure both after optimization, but do not use the structure metrics to select prompts.
\item \textbf{The place value hurdle and how to clear it.} In the harder setting, the base system struggles to retain an injective sender code and gives Grace too few confirmed examples. A collision penalty for Rocky, a window that keeps confirmed successes for Grace, and optimizing Rocky before Grace can enable a generalizing place value code.
\end{itemize}

\stepcounter{footnote}\footnotetext{Character art throughout this paper is OpenMoji,
    \url{https://openmoji.org}, CC BY-SA 4.0.}

% ---- sections/02-related-work.typ ----
\section{Related work}\label{sec:related}

Emergent communication with frozen large language models is a highly
interdisciplinary problem. Cognitive science and linguistics ask when
languages emerge and how compositionality arises. The emergent communication
literature has created a set of toy games to test that specific problem.
More
recently, a line of work asks if the simple toy agents used in emergent
communication can be replaced by much more capable Large Language Models.
Finally, prompt optimization updates the frozen models' prompts from scored interactions without changing their weights.

\paragraph{Language emergence and compositionality.}
How does compositionality arise in a language? \cite{kirby2008cumulative} answer
that question via iterated learning. Here, a language is passed through a
chain of learners, each seeing only a fragment, and becomes compositional
because structure is all a learner can rebuild from a fragment, so structure
is all that survives the handoff. The later work \cite{kirby2015compression}
explains why. A language is squeezed between being compressible, so it can
be learned from little, and expressive, so it can distinguish everything,
and compositional structure is the equilibrium. We apply a related squeeze through a memory window too small for the full codebook, while optimized genes persist across rounds. Our setup does not transmit a language through generations of fresh learners. \cite{lazaridou2017emergence}
brought this line of work to deep learning where two trained networks
invented a code in a referential game. Two results from that line anchor how
we measure.

\cite{chaabouni2020compositionality} find no correlation between compositionality
and generalization, but a more compositional language is picked up more
easily by fresh listeners, even ones with a different architecture. Their
disentanglement metrics are what we use to analyze the language
obtained. \cite{ren2020compositional} build the bottleneck into training itself by
using a chain of networks whose weak pre-training favors compositional
languages each generation. Here structure is selected for by construction
rather than observed. Both lines run small recurrent agents rather than
language models: gated recurrent units in \cite{chaabouni2020compositionality},
and an LSTM speaker paired with an LSTM listener in \cite{ren2020compositional}.
The two studies report different relationships between structure and generalization in their settings. We choose to sit deliberately in the small space regime. Both of the above works train their agents from scratch and cannot tell us what they would do if the weights stay frozen.

\paragraph{Referential and reconstruction games.}
The choice of the game that the agents play is critically important as depending on the level of supervision and some of the key design parameters of the game, the agents can sometimes learn incredibly well or not. The reconstruction game from \cite{chaabouni2020compositionality} works in their paper because their Grace is supervised almost like the decoder layer of an autoencoder. The object that is being reconstructed has a rich inner activation and as a result has enough supervisory signal to be able to allow for emergent communication. A referential game is slightly different and is the more general, but slightly easier form of the reconstruction. Instead, Grace is given a candidate set (not the entire object space however) and is given a binary success signal if he was able to successfully identify the right object out of that set. More formally, reconstruction and referential games are two ends of a spectrum, if we assume that the number of candidates is \(N = W^{K}\), the referential game becomes the reconstruction game and depending on the value of \(N\) the difficulty of the game varies.

\paragraph{Emergent communication with frozen LLMs.}
Our work is heavily inspired by the work done in emergent communication from frozen LLMs. The earliest work, \cite{kouwenhoven2024searching}, uses one frozen Llama-3-70B model to play both roles in a referential game. Unlike our case, they allow the messages to grow in length and do not have a fixed message length as is standard in referential games. Consequently, many objects have the same signal which produces a non-injective code. \cite{talebirad2026signals} also uses one frozen LLM to play both Rocky and Grace in a referential game but their main thesis is comparing five different hand designed memory harnesses. Their agents usually use lookup tables and they show that the memory architecture changes the emergent codes. Both of these studies use the same model for both players, whereas we use different ones for our Rocky and Grace.

Recently, \cite{beltoft2026emergent} look further out by performing ethology on an
open population of agents on a social platform, where agents propose new
languages ranging from token-efficient codes to codes built to evade
oversight. Unlike ours, this is closer to fieldwork than to a controlled
experiment. Instead of observing agents in the wild, we create a controlled
environment and experiment to allow the agents to develop their own
communication strategies.

\paragraph{Prompt optimization.}
In our frozen setting, prompt optimization is the mechanism for adapting the agents' prompts. A family of optimizers
rewrites a program's prompts against a metric: search by an LLM optimizer
\cite{yang2023llmoptimizers}, textual gradients (\cite{pryzant2023protegi},
\cite{yuksekgonul2024textgrad}), and joint instruction-demonstration search in the
DSPy line (\cite{khattab2024dspy}, \cite{opsahlong2024optimizing}). We use GEPA \cite{agrawal2025gepa}, a reflective evolutionary prompt optimization strategy. It reflects by having a surrogate reflection model that proposes new prompts based on scored rollouts. GEPA evaluates candidate prompts and allows them to replace the old prompt if they improve their score on a validation set. A parallel line of work optimizes the compound agent architecture itself (\cite{zhuge2024gptswarm}, \cite{hu2024adas}, \cite{wu2025optimas}) but in our setting we hold the architecture constant while the agents' genes and memory windows are allowed to change.

To contextualize our work with all of the related work, we use two different frozen LLMs, accessible only via an API to communicate through a constrained message channel and optimized only via prompt optimization. We compute the structure of the communication code only after optimization. To make it easier we use Table~\ref{tab:positioning} to compare our design choices with the four closest prior work. Different players means that Rocky and Grace have fundamentally different model backends. Frozen weights means the models are not updated. Fixed architecture means that the agent design doesn't vary across the compared examples.

\begin{table}[H]
\caption{\textbf{Design choices in related communication studies.} The columns compare frozen player weights, different model backbones for the two roles, and fixed agent architecture.}\label{tab:positioning}
\centering
\begin{tabulary}{\linewidth}{LL>{\centering\arraybackslash}p{2.0cm}>{\centering\arraybackslash}p{2.2cm}>{\centering\arraybackslash}p{2.4cm}}
\toprule
 & agents & frozen weights & different players & fixed \newline architecture \\
\midrule
\cite{chaabouni2020compositionality} & trained GRUs & \no{} & \no{} & \yes{} \\
\cite{ren2020compositional} & iterated LSTMs & \no{} & \no{} & \yes{} \\
\cite{kouwenhoven2024searching} & 1 frozen LLM & \yes{} & \no{} & \yes{} \\
\cite{talebirad2026signals} & 1 frozen LLM & \yes{} & \no{} & \no{} \\
\textbf{Our System} & \textbf{2 frozen LLMs} & \yes{} & \yes{} & \yes{} \\
\bottomrule
\end{tabulary}
\end{table}

% ---- sections/03-method.typ ----
\section{Methods}\label{sec:method}

We need four pieces to put together our frozen LLM agents to talk to each
other. We need to define the game (Section~\ref{sec:game}) that the agents play with each
other, and in this setting one message is sent from the sender, Rocky, to
the receiver, Grace, and one reward bit says whether Grace picked the object
Rocky was shown. This game is a general framework and there are two
specific settings we use to study positional and place value codes (Section~\ref{sec:sizings}). Once we
establish the game, we need to define the anatomy of each agent
(Section~\ref{sec:agents}): its two modules, their genes and harness, and its memory
window. As we access the frozen LLMs through API endpoints, the design choices in this section are critical to ensure that our
agents are indeed able to play this game (Section~\ref{sec:agents}). Next, we need to
define how we optimize each agent, here via the Prompt Optimization
procedure and the memory window such that the agents are indeed able to learn over time
(Section~\ref{sec:optimization}). Finally, we describe the prompts and the harness behind
each agent, the fixed text that no optimizer may touch, along with the
specific open frozen LLM models we use.

\subsection{The referential game}\label{sec:game}

An object is a string of \(k\) letters, each drawn from an object alphabet of
\(W\) values, so the object space holds \(W^{k}\) objects. A message is a string
of exactly \(l\) symbols from a message alphabet of \(V\) symbols; the
channel admits \(V^{l}\) distinct messages, and the harness refuses any sizing
with \(V^{l} < W^{k}\). This means that an injective mapping exists from the object space into the message space. Such a map is invertible on its image, and we call it a code.

\begin{figure}[H]
\centering
\includegraphics[scale=0.6]{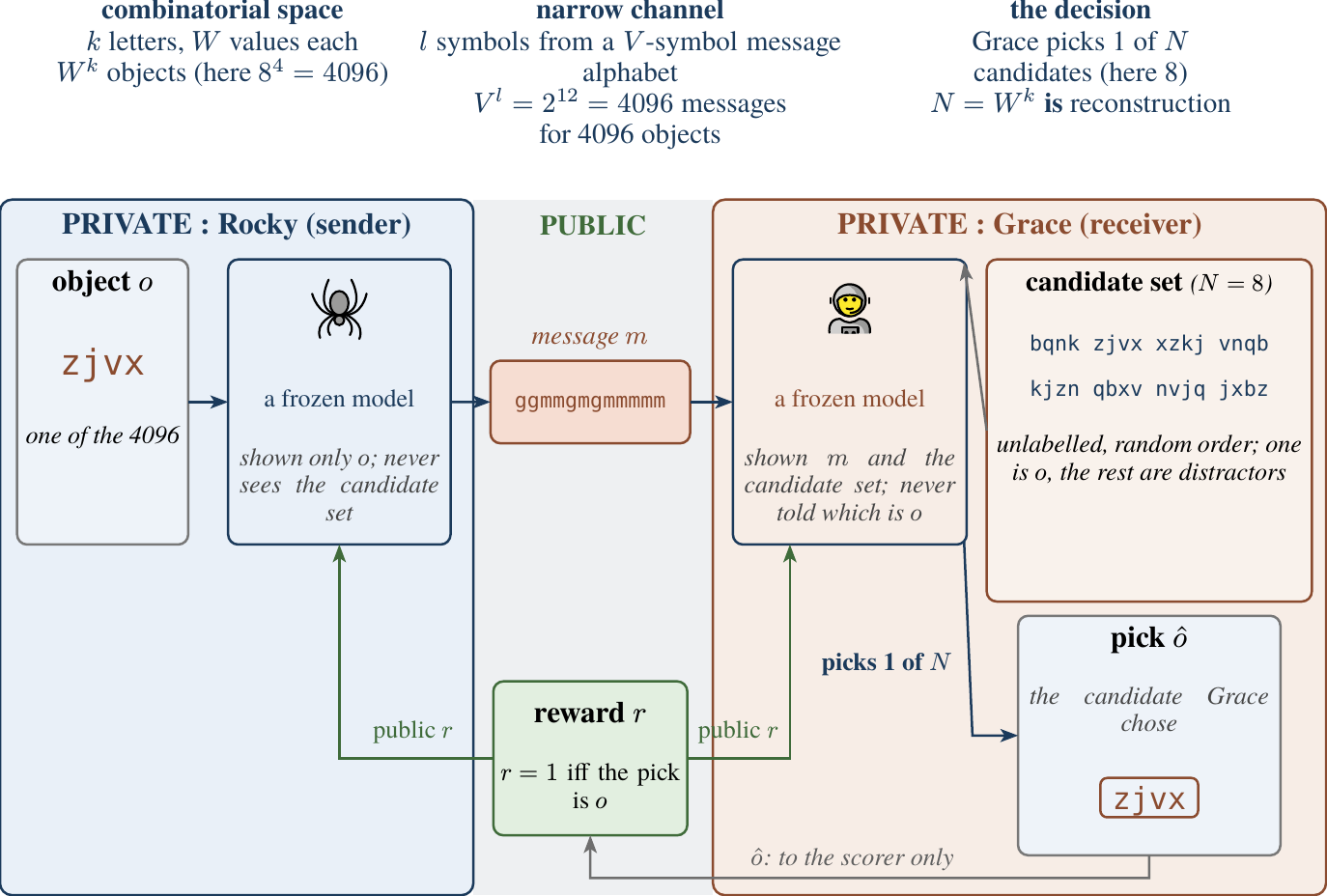}
\caption{\textbf{One round of the game.} Rocky sees the object and emits a message over
    the message alphabet. Grace sees the message and an unlabelled
    candidate set and picks one from it. Rocky sends \(m\) and the environment returns \(r\) to both agents.}\label{fig:game-anatomy}
\end{figure}

Both the object alphabet and the message alphabet are scrambled, opaque,
single-token symbols, fixed by a token seed, and the object alphabet and
message alphabet share no symbols. The disjoint scrambled alphabets prevent direct copying of the object into the message; no object-to-message mapping is supplied in the seed genes. Each round, as drawn in
Figure~\ref{fig:game-anatomy}, the sender, Rocky, sees one object \(o\), drawn at random,
and emits one message \(m\). The receiver, Grace, sees \(m\) and a candidate set
\(\mathcal{C}\) of \(N\) objects with \(o\) among them, and picks one, \(\hat{o}\).

The environment compares Grace's pick with the target and returns the reward \(r\) to both agents: 1 when \(\hat{o} = o\) and 0 otherwise. Rocky's message and the reward are shared, while the target object and candidate set remain private to their respective agents. Neither agent sees the other's committed rule.

Table~\ref{tab:lexicon} describes the jargon we use to describe the referential game. The specific instantiation settings of the referential game can be seen in Table~\ref{tab:notation}.

\begin{table}[H]
\caption{\textbf{The lexicon of the game.} One word and one symbol per part, used
    unchanged in every section, figure and formula. The sizes are in
    Table~\ref{tab:notation} and the agents' parts in Table~\ref{tab:agent-lexicon}.}\label{tab:lexicon}
\centering
{\tablesize
\begin{tabularx}{\linewidth}{ll>{\raggedright\arraybackslash}Xl}
\toprule
term & symbol & meaning & in the example \\
\midrule
object alphabet & \(\mathcal{L}\), \(|\mathcal{L}| = W\) & the \(W\) values a letter can take & \code{b z n v k j q x} \\
object & \(o \in \mathcal{L}^{k}\) & a string of \(k\) letters, one of \(W^{k}\) objects & \code{zjvx} \\
letter & \(o_{i} \in \mathcal{L}\) & the \(i\)-th letter of an object & \(o_{2} =\) \code{j} \\
message alphabet & \(\mathcal{S}\), \(|\mathcal{S}| = V\) & the \(V\) symbols a message is written in & \code{g m} \\
message & \(m \in \mathcal{S}^{l}\) & a string of \(l\) symbols, one of \(V^{l}\) messages & \code{ggmmgmgmmmmm} \\
symbol & \(m_{j} \in \mathcal{S}\) & the \(j\)-th symbol of a message & \(m_{4} =\) \code{m} \\
candidate set & \(\mathcal{C}\), \(|\mathcal{C}| = N\) & the \(N\) objects Grace is shown, \(o\) among them & \code{bqnk zjvx} ... (8) \\
pick & \(\hat{o} \in \mathcal{C}\) & the candidate Grace chose & \code{zjvx} \\
reward & \(r \in \left\{ 0,1 \right\}\) & 1 if \(\hat{o} = o\), else 0 & 1 \\
game &  & the \posgame{} or the \pvgame{}, the two sizings of Table~\ref{tab:notation} & the \posgame{} \\
memory window & \(D\) & the last \(D\) rounds an agent's prompt shows it (Section~\ref{sec:agents}) & 19 rounds \\
observed objects & \(\mathcal{O}\) & the objects a run played & 462 at 4096 objects \\
code & \(c:\mathcal{O} \rightarrow \mathcal{S}^{l}\) & the message a run sent most often for each object it played & \(c\left( \texttt{zjvx} \right) =\) \code{ggmmgmgmmmmm} \\
\bottomrule
\end{tabularx}}
\end{table}

This game is formally known as the referential game in the literature
\cite{lazaridou2017emergence} and is a generalization of the reconstruction game
of \cite{chaabouni2020compositionality}. In the reconstruction game, as the name
suggests, Grace must recover the target object from the message alone, with
no candidate set to narrow it down: in effect he sees all \(W^{k}\) objects and
must reconstruct the target. In our referential game, he sees only \(N\)
candidates and picks the target object from that set. As mentioned before, the reconstruction and referential games are at two opposite ends of a spectrum of difficulty. As \(N\) grows, the referential game approaches the reconstruction game and becomes more difficult.

\subsection{The two sizings}\label{sec:sizings}

In this paper, we use two specific instantiations of the referential game.
The two specific settings are given in Table~\ref{tab:notation}. The first game
utilizes a \textbf{positional code} which essentially allows for each letter to
be given its own fixed block of message positions. Each letter in the object
space is given its own per letter or block code as the literature suggests
\cite{chaabouni2020compositionality}. In our paper, we refer to this code as a
positional code henceforth. As \(k = 4\), \(l = 12\), \(V = 2\) and \(W = 8\), each letter
of the object needs exactly \(\left\lceil {\log_{2}8} \right\rceil = 3\) bits in the message space
and four letters is exactly the 12 bits that the channel allows.

\begin{figure}[H]
\centering
\includegraphics[scale=0.88]{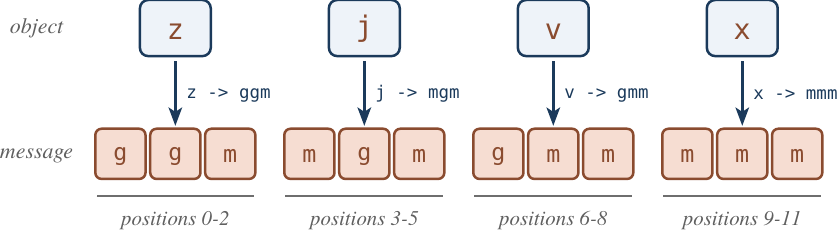}
\caption{\textbf{The positional code on a real emitted pair.} Each letter corresponds to 3
    fixed message positions, so one confirmed pair can reveal up to \(k\) distinct letter facts. The message alphabet varies per random seed.}\label{fig:block-code}
\end{figure}

One instantiation of the positional code can be seen in Figure~\ref{fig:block-code},
where the object \code{zjvx} is encoded as \code{ggm mgm gmm mmm}. Each letter of the
object is given its own fixed block of message positions. Under this positional interpretation, a reward of one makes up to four distinct letter facts available to Grace; repeated letters yield fewer distinct facts. In this case, Grace is
able to see that \code{z} maps to \code{ggm}, \code{j} maps to \code{mgm}, \code{v} maps to \code{gmm} and
\code{x} maps to \code{mmm}.

\begin{table}[H]
\caption{\textbf{The two games.} One per code family, at the values every number in the
    paper uses. The \posgame{} admits a positional code exactly; the \pvgame{}
    has no room for three independent fixed letter blocks.}\label{tab:notation}
\centering
\begin{tabulary}{\linewidth}{LLCC}
\toprule
symbol & meaning & \posgame{} (positional) & \pvgame{} (place value) \\
\midrule
\(k\) & letters per object & 4 & 3 \\
\(W\) & object alphabet size & 8 & 6 \\
\(W^{k}\) & objects & 4096 & 216 \\
\(l\) & symbols per message & 12 & 8 \\
\(V\) & message alphabet size & 2 & 2 \\
\(N\) & candidates shown & 8 & 20 \\
\(D\) & memory window rows & 19 & 19 \\
\bottomrule
\end{tabulary}
\end{table}

However, the second game does not fit three independent letter blocks. When
\(k = 3\), \(W = 6\), \(V = 2\) and \(l = 8\), each letter needs
\(\left\lceil {\log_{2}6} \right\rceil = 3\) bits per block and consequently three letters require nine
bits yet the channel only has eight. One possible solution could be a variable-length code for each letter but this does not solve the problem either. Because objects can repeat a word up to 3 times, a letter codeword cannot be longer than two symbols. Unfortunately, a binary code of that length can only have four codewords which is not enough for six letter values. However, an injective code does exist as there are 216 objects and a total of \(2^{8} = 256\) messages so there is room to spare. One systematic and intuitive solution is a base-6 place value code that the receiver will have to compute rather than a block to letter table. The algorithm essentially is to read the object's three letters as the digits of a base-6 number and then write that number in binary in the message space for the channel.

\begin{figure}[H]
\centering
\includegraphics[scale=0.9]{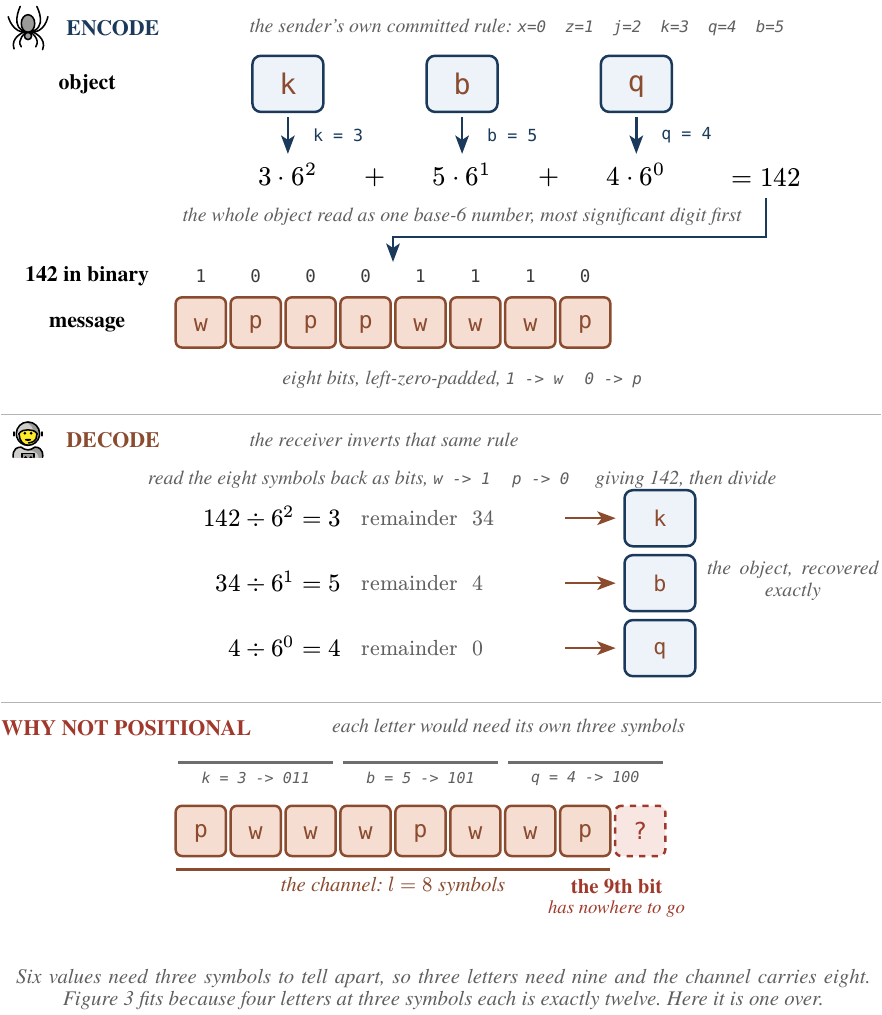}
\caption{\textbf{The place value code and why a positional code cannot work} Top, the pair's
    own rule, read from the settled Rocky's committed rule where each letter is a
    base-6 digit, the three digits are one number, and that number is
    written as eight bits. Bottom, a positional code that needs nine bits for
    three letters yet the channel has only eight.}\label{fig:place-value}
\end{figure}

To make this more concrete, we show an example of the place value code in Figure~\ref{fig:place-value}. The code in the reported successful \pvgame{} runs maps digits as \code{x=0, z=1, j=2, k=3, q=4, b=5}, then reads the whole object as a base 6 number and then writes that as a binary number in 8 bits. Here \code{wpppwwwp} is 142. To reverse the computation, Grace writes \(142 = 3 \cdot 36 + 5 \cdot 6 + 4\) and then reads the digits as \code{kbq}. The key difference between the positional code and the place value code is that there is no partial credit available to Grace. Grace cannot learn one letter at a time as a confirmed pair constrains the whole rule rather than individual letter blocks. The additional difficulty of the \pvgame{} is what Experiment 2 measures (Section~\ref{sec:boundary}).

\subsection{The agents}\label{sec:agents}

\begin{figure}[H]
\centering
\includegraphics[scale=0.92]{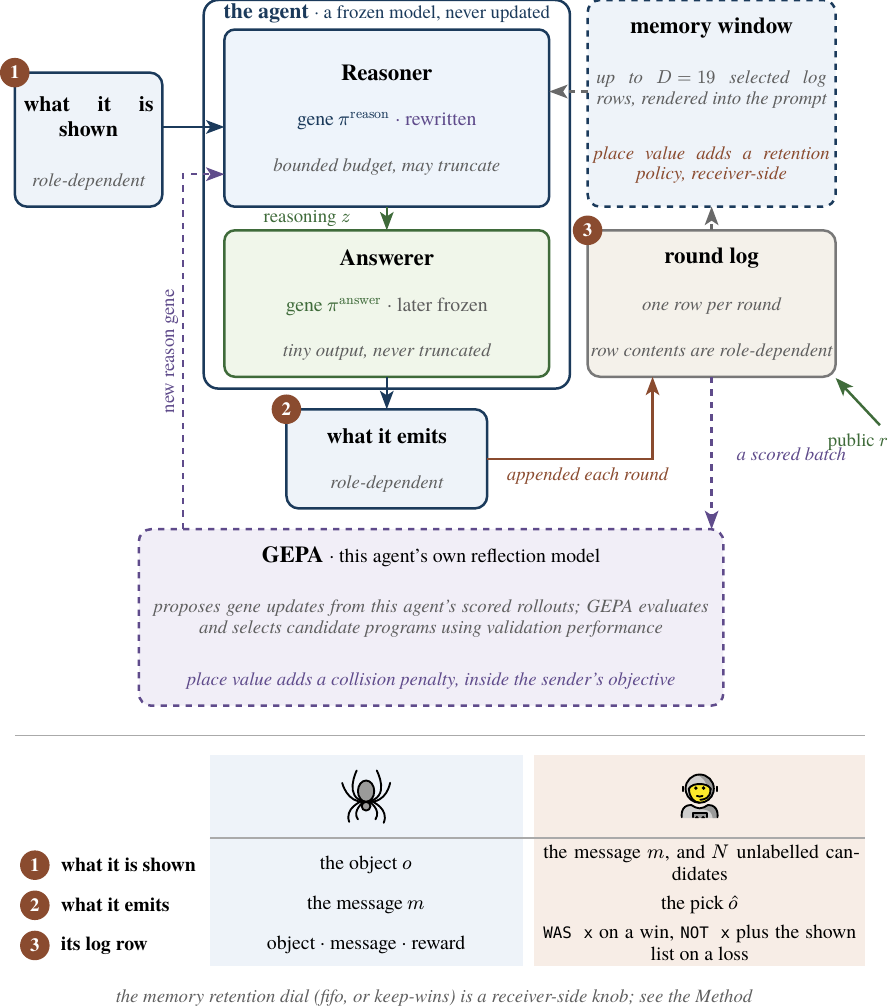}
\caption{\textbf{Inside an agent.} Rocky and Grace share the same anatomy which contains a bounded
    reasoner, a tiny answerer, a round log with up to \(D\) selected rows injected into the
    memory window, and the agent's own GEPA job. The dashed boxes mark the optimizer and window, and the numbered slots are role dependent.}\label{fig:agent-anatomy}
\end{figure}

We define the term \textbf{gene} to refer to a single optimizable instruction text. These are the instructions that a module or a prompt in layman's terms runs and is the only part of the system that changes. A module is defined by a DSPy typed signature \cite{khattab2024dspy} so a gene is that typed signature's instruction text. We define everything else in the module that we cannot optimize as the \textbf{harness} (Section~\ref{sec:harness}). Each agent is comprised of two modules carrying one gene each (see Figure~\ref{fig:agent-anatomy}) for more details. Therefore, there are four genes in total: Rocky's \(\pi_{\text{Rocky}}^{\text{reason}}\) and \(\pi_{\text{Rocky}}^{\text{answer}}\), and Grace's \(\pi_{\text{Grace}}^{\text{reason}}\) and \(\pi_{\text{Grace}}^{\text{answer}}\). The specific jargon we use to define and describe agents is given in Table~\ref{tab:agent-lexicon}.

\begin{table}[H]
\caption{\textbf{The agent lexicon.} One word per part of an agent, per step of changing
    one, and per unit of running the pair. The game's own parts are in
    Table~\ref{tab:lexicon}.}\label{tab:agent-lexicon}
\centering
{\tablesize
\begin{tabularx}{\linewidth}{l>{\raggedright\arraybackslash}X}
\toprule
term & meaning \\
\midrule
agent & Rocky, the sender, and Grace, the receiver \\
module & one typed DSPy signature an agent runs. The \textbf{reasoner} thinks under a token budget and \textbf{answerer} formats the answer \\
gene & one module's instruction text and the only text an optimizer may rewrite \\
seed gene & the original gene a run starts from (Section~\ref{sec:harness}) \\
optimized gene & a gene after the optimizer has rewritten it \\
harness & the fixed text around a gene, which no optimizer may touch (Section~\ref{sec:harness}) \\
committed rule & the rule the reasoner writes down each round and is re-injected into its prompt the next round \\
round log & the agent's own record of the rounds it played with one rollout per round \\
memory window & the last \(D\) rows from the round log and is chosen by the retention policy and injected into the prompt \\
\midrule
optimizer & Prompt Optimizer (here, GEPA) \\
reflection model & the model the optimizer calls to propose a rewritten gene \\
firing & one GEPA invocation for one agent but it may evaluate multiple prompt candidates \\
validation set & the fixed objects a priori used to evaluate candidate programs during prompt optimization \\
\midrule
the pair & The agents Rocky and Grace \\
round & One iteration of play. Rocky sees \(o\) and sends \(m\); Grace sees \(m\) and \(\mathcal{C}\) and picks \(\hat{o}\); both receive \(r\) (Figure~\ref{fig:game-anatomy}) \\
rollout & The full record of one round: what each module read and produced, the pick, and \(r\); the round log and the optimizer's reflection minibatch \\
budget & the number of rounds a run may play and optimization is included \\
arm & a way of spending the round budget, named by its cell of the 2x2 grid. \code{no-prompt-opt@mem} plays with the memory window on and no optimizer, \code{prompt-opt@nomem} gives every round to the optimizer with the window off, \code{prompt-opt@mem} runs the optimizer with the window on, and \code{no-prompt-opt@nomem}, the seed genes with the window off, needs no run (Section~\ref{sec:optimization}) \\
run & one arm at one sizing and one seed \\
held-out objects & objects used to determine accuracy but are never seen during a run \\
\bottomrule
\end{tabularx}}
\end{table}

We split these agents into two modules because under a capped budget of
tokens (necessary for practical experiments), the answer is often truncated
or missing if a single module is used. Depending on the reasoning model and
the current state of the system, the reasoning may be truncated mid-thought
and the answer may not be present in the output. This is especially true for
longer contexts where the reasoning model needs to maintain a long
chain of thought. By splitting the agent into two modules, even if the
reasoning model is truncated mid-thought, the answerer module can still
produce a valid answer based on the reasoning instead of malformed outputs.
The answerer is essentially just a short output field, small enough that
truncation can never reach it.

The two agents only differ in the information that passes through their identical anatomy. Rocky's reasoner is shown the target object and his reasoner emits the message. Grace's reasoner meanwhile sees the message and the unlabelled candidate set and his answerer picks one of the candidates. The memory windows also differ. Each row of Rocky's window is object, message, reward, while Grace's window is message, candidate set, pick, reward with \code{WAS X} on a win and \code{NOT X} otherwise.

Each agent's round log and memory (if it is needed) is stored and injected into its prompt. For this paper, where \(D = 19\), the window covers less than 9\% of the smallest object space and cannot hold a lookup table of message-object pairs. At depth zero, the window renders as \code{(no past interactions yet)}. When the memory is off, the memory window is not injected in. The retention policy varies between the default, first in first out or fifo and keep-wins which first evicts losing rollouts before winning ones.

\subsection{Prompt optimization}\label{sec:optimization}

We use prompt optimization, a structured mechanism to improve the agent's genes over time to allow for communication between agents. We specifically use GEPA \cite{agrawal2025gepa}, which is a genetic algorithm that uses a surrogate reflection model to update genes from batches of rollouts. It evaluates candidate genes and selects one that performs best on the validation sets. Note that each agent optimizes its own genes alone without the other seeing it at all. Both agents, by default, call the same reflection model. Note that only the message and the reward are seen by both agents and everything else is individual to each of them. The optimizer is not part of the referential game at all and in our experiments we use the DSPy implementation of GEPA.

For the positional code experiments, both the reasoner and the answerer genes are optimizable but for the place value code only the reasoner is. The reason for this is because GEPA injected a codebook into the answerer gene which is against the spirit of the two stage design. Therefore, to preserve the design, we froze the answerer genes from the third place value attempt onwards. The seed genes contain no explicit codebook, but an optimized gene can. We identify results that rely on such a gene and reproduce it in Section~\ref{sec:app-genes}.

Every single experiment has a fixed round budget from which both rounds and GEPA scoring are drawn. We have multiple arms or methods of spending that budget and there are a total of four with each in a 2x2 grid in Table~\ref{tab:knobs}. Each cell is labelled as \code{prompt@window} with \code{no-prompt-opt} for using the seed genes the whole time and \code{prompt-opt} for using the optimizer to update the genes. For the window, \code{mem} indicates using the memory window on while \code{nomem} is intuitively the memory window off. If we use the alternate retention policy keepwins, we use \code{memwins}. Note that the \code{prompt-opt@mem} has a schedule that splits its rounds between play and optimizer firings so there are less rounds of pure play.

Evaluation of all the arms are done on the held out objects that are never seen in play or optimization. To evaluate the effect of the prompt in \code{prompt-opt@mem}, we re-evaluate the optimized gene without memory and we denote that as \code{prompt-opt-both@nomem}. We do use repeated draws to measure variation in the evaluation responses. Note that this is not variation across optimization runs. During evaluation, the genes are fixed and the memory does not change either. Note that the validation set used for prompt selection in optimization is separate from the final held out testing set for evaluation all of the examples. Further details are given in Section~\ref{sec:app-config-eval}.

\begin{table}[H]
\caption{\textbf{The two knobs and the five cells.} Memory on or off or prompt optimization on or off, with
    \(D = 19\). Each of
    the four cells is one arm and the fifth reuses the \code{prompt-opt@mem}'s optimized genes with the memory removed.}\label{tab:knobs}
\centering
\begin{tabulary}{\linewidth}{LCC}
\toprule
 & \textbf{memory off} (\code{nomem}) & \textbf{memory on} (\code{mem}) \\
\midrule
\textbf{seed genes} & \code{no-prompt-opt@nomem} \newline neither knob & \code{no-prompt-opt@mem} \newline memory alone \\
\textbf{optimized genes} & \code{prompt-opt@nomem} \newline the optimizer alone & \code{prompt-opt@mem} \newline optimizer and memory \\
\midrule
\textbf{re-evaluation} & \multicolumn{2}{c}{\cellstack{c}{\code{prompt-opt-both@nomem} \\ \code{prompt-opt@mem}'s genes without the memory}} \\
\bottomrule
\end{tabulary}
\end{table}

\subsection{The harness around the genes}\label{sec:harness}

\begin{figure}[H]
\centering
\includegraphics[scale=0.88]{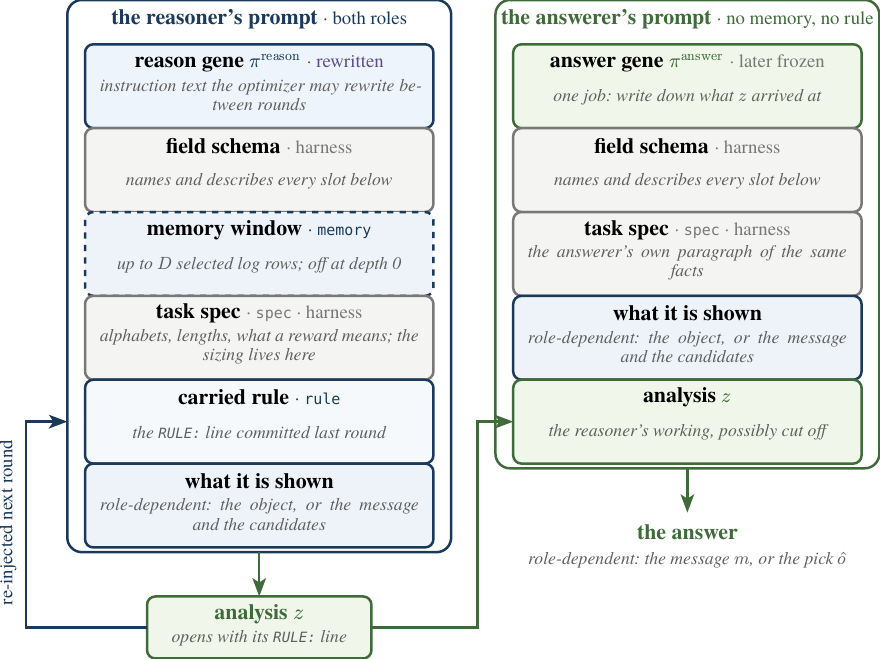}
\caption{\textbf{The two harnesses in each agent} Each module consists of both its gene and its harness (in grey). The reasoner's prompt
    carries the two slots with state, the memory window and last round's
    rule while the answerer's prompt merely carries the analysis.}\label{fig:prompt-assembly}
\end{figure}

Everything that an agent's frozen LLM takes in as input other than its gene is referred to as the \textbf{harness}. The harness is fixed text that the optimizer is not allowed to touch. Figure~\ref{fig:prompt-assembly} shows the full architecture of the input to
the two agents and the harness is clearly displayed.

\subsubsection{Rocky's harness and seed genes}\label{sec:seed-sender}

Rocky's harness specifies the game for both modules and gives the reasoner the memory window and rule carried over. We also attach his seed reasoner and answerer genes.

\begin{harnessbox}{Rocky's; \pvgame{}}
\begin{Verbatim}[fontsize=\fontsize{7.20pt}{6.88pt}\selectfont]
### spec/sender.reason
Objects are exactly 3 characters long, and every character is one of these 6: jkqvbz. Messages are exactly 8 characters long, and every character is one of these 2: rd. The message you settle on must be exactly 8 message characters, with no spaces. A partner who sees only your message must reconstruct the object. In your history, reward 1 means the object was reconstructed correctly, reward 0 means it was not.

### spec/sender.answer
Objects are exactly 3 characters long, and every character is one of these 6: jkqvbz. Messages are exactly 8 characters long, and every character is one of these 2: rd. Your message field must contain exactly 8 message characters and nothing else, with no spaces. A partner who sees only your message must reconstruct the object. In your history, reward 1 means the object was reconstructed correctly, reward 0 means it was not.

### rule/RULE_FIELD_DESC
the rule you committed to on your last turn

### rule/NO_RULE
(no rule committed yet -- this is your first turn)

### window/NO_MEMORY
(no past interactions yet)
\end{Verbatim}
\end{harnessbox}

\begin{sendergene}{seed}
\begin{Verbatim}[fontsize=\fontsize{7.20pt}{6.88pt}\selectfont]
You are the sender in a communication game. You see an OBJECT (a short string of
object characters) and must decide on a MESSAGE (a fixed-length string of message
characters). Your partner sees only your message and must write down the object.
You are scored 1 when their answer matches the object exactly, and 0 otherwise.

Do not write the message down here. Work the encoding out, say what rule you are
using, and apply it to this object.

Begin your analysis with a single line of the form
RULE: <the rule you are using, stated compactly enough to reuse verbatim>
and only then do the working for this turn.

You are given the rule you committed to on your last turn. Restate it UNCHANGED
unless a row in your memory contradicts it. Do not replace a working rule with a new one
just because the last turn
scored 0. If you do revise it, say on the RULE line what changed and why.
\end{Verbatim}
\end{sendergene}

\begin{senderanswer}{seed; frozen from the third place value attempt on}
\begin{Verbatim}[fontsize=\fontsize{7.20pt}{6.88pt}\selectfont]
You are the sender in a communication game. The analysis below is the working for
this object, already done. Write down the MESSAGE it arrives at, in message
characters, and nothing else. The analysis may stop mid-sentence; if it does, take
the message it was best supporting. Do not redo the working and do not explain.
\end{Verbatim}
\end{senderanswer}

\subsubsection{Grace's harness and seed genes}\label{sec:seed-receiver}

Grace's harness adds the candidate set and outcomes to the game specification. We also attach his seed reasoner and answerer genes.

\begin{harnessbox}{Grace's; \pvgame{}}
\begin{Verbatim}[fontsize=\fontsize{7.20pt}{6.88pt}\selectfont]
### spec/receiver.reason
Messages are exactly 8 characters long, and every character is one of these 2: rd. Objects are exactly 3 characters long, and every character is one of these 6: qzbvkj. You will also be shown a numbered list of 20 candidate objects, exactly one of which is the object the sender meant. Work out which of the 20 candidates the sender meant -- the answer is always one of them, never an object outside the list (refer to it by its 3 object characters or its number). In your history, reward 1 means you picked the correct object, reward 0 means you did not. Each history row also lists the candidates you were shown that turn.

### spec/receiver.answer
Messages are exactly 8 characters long, and every character is one of these 2: rd. Objects are exactly 3 characters long, and every character is one of these 6: qzbvkj. You will also be shown a numbered list of 20 candidate objects, exactly one of which is the object the sender meant. Your answer must be ONE of those candidates -- choose it and put its 3 object characters in your guess field, with no spaces (you may instead give its number). Do not output an object that is not in the list. In your history, reward 1 means you picked the correct object, reward 0 means you did not. Each history row also lists the candidates you were shown that turn.

### window/OUTCOME_LEGEND
(WAS x: the object was x.  NOT x -> one of: a b c: not x, but one of those.)
\end{Verbatim}
\end{harnessbox}

\begin{receivergene}{seed}
\begin{Verbatim}[fontsize=\fontsize{7.20pt}{6.88pt}\selectfont]
You are the receiver in a communication game. You see a MESSAGE and a numbered list of
CANDIDATE objects, exactly one of which is the object the sender meant. Decide WHICH
candidate -- this is a choice among the listed objects, and the answer is always one of
them, never an object outside the list. You are scored 1 when you choose the sender's
object and 0 otherwise.

Say which numbered candidate you choose.
Do not settle on an object that is not in the list.

Begin your analysis with a single line of the form
RULE: <the rule you are using, stated compactly enough to reuse verbatim>
and only then do the working for this turn.

You are given the rule you committed to on your last turn. You are the one reading, so a 0 is
never your partner's misreading: it means the object was not the candidate you
chose. The message itself may or may not have followed your partner's rule.
\end{Verbatim}
\end{receivergene}

\begin{receiveranswer}{seed; frozen from the third place value attempt on}
\begin{Verbatim}[fontsize=\fontsize{7.20pt}{6.88pt}\selectfont]
You are the receiver in a communication game. Below is a numbered list of CANDIDATE
objects and an analysis of which one the sender meant. Your only job is to COMMIT to one
of the listed candidates.

Answer with the candidate's NUMBER. Every reply must be one of those numbers -- there is
always an answer, and refusing or leaving it blank scores the same as being wrong.

The analysis is often cut off before it reaches a conclusion. That is expected and is not
a reason to withhold an answer: pick the candidate it leans towards. Objects mentioned in the
analysis or in past rows are NOT answers unless they appear in this rollout's candidate
list. Do not redo the working.
\end{Verbatim}
\end{receiveranswer}

\subsubsection{The Reflection Model's harness}

The reflection model's harness is very similar to the agents' harnesses. The two key changes are that the incumbent gene and the minibatch of scored rollouts with that particular agents inputs and response are also encoded. For example, Rocky sees \code{reward 1: object zjvx -> message \textquotesingle{}ggmmgmgmmmmm\textquotesingle{} was identified by your partner.} The reflection model returns a candidate gene for GEPA to evaluate.

\begin{figure}[H]
\centering
\includegraphics[scale=0.88]{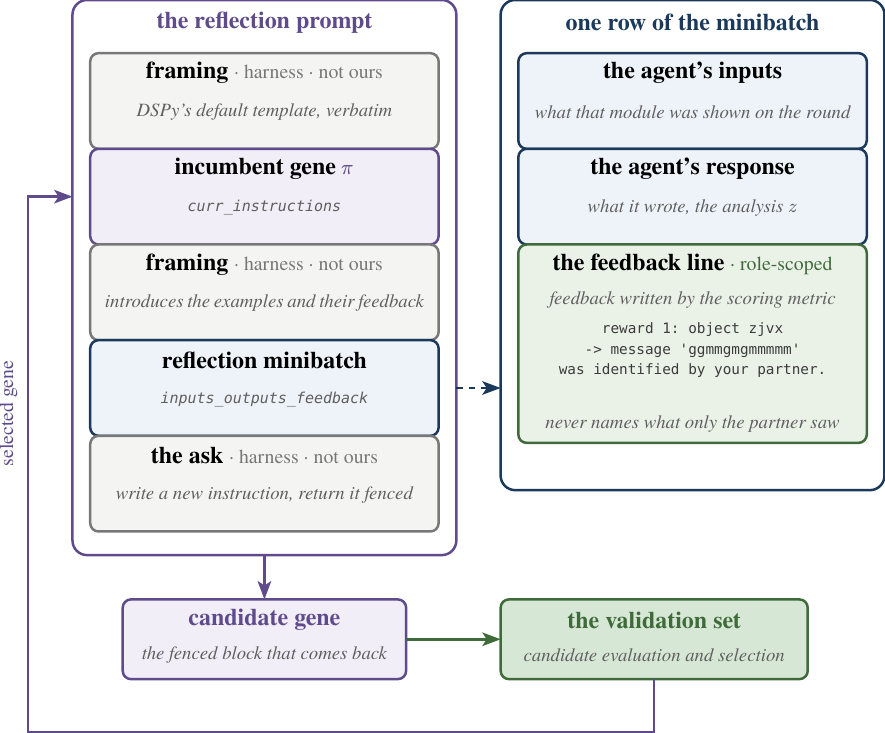}
\caption{\textbf{The optimizer's harness.} Left, the reflection prompt, DSPy's fixed
    text with the two coloured slots this system fills. Right, one scored
    rollout in the minibatch slot. Bottom, GEPA evaluates the returned candidates through its validation procedure.}\label{fig:optimizer-harness}
\end{figure}

\begin{reflectionbox}{the optimizer package's default template}
\begin{Verbatim}[fontsize=\fontsize{7.20pt}{6.88pt}\selectfont]
I provided an assistant with the following instructions to perform a task for me:
```
<curr_instructions>
```

The following are examples of different task inputs provided to the assistant along with the assistant's response for each of them, and some feedback on how the assistant's response could be better:
```
<inputs_outputs_feedback>
```

Your task is to write a new instruction for the assistant.

Read the inputs carefully and identify the input format and infer detailed task description about the task I wish to solve with the assistant.

Read all the assistant responses and the corresponding feedback. Identify all niche and domain specific factual information about the task and include it in the instruction, as a lot of it may not be available to the assistant in the future. The assistant may have utilized a generalizable strategy to solve the task, if so, include that in the instruction as well.

Provide the new instructions within ``` blocks.
\end{Verbatim}
\end{reflectionbox}

\subsection{Models}\label{sec:models}

For our experiments, we use two distinct open weight LLMs. Rocky, the sender, is GLM-5 \cite{zeng2026glm5}, and Grace, the receiver, is DeepSeek-V4-Flash \cite{deepseek2026v4}. They are post-trained models from different providers, accessed through API endpoints, with different pretraining, tokenizers, architecture and post-training. These choices are deliberate: LLMs recognize and favor their own generations \cite{panickssery2024selfrecognition}, so a code formed between two copies of one model may benefit from shared weights. Our procedure does not update either model's weights, decoding settings, or architecture; the input prompts and retained state can change.

We picked our agents' model for a clear and specific reason. The first is
cost. Every experiment requires hundreds of rounds between Rocky and Grace, so the models must be strong enough
to reason but cheap enough to run at scale. Secondly, we need a reasoning trace. Rocky and Grace's reasoning modules must return enough reasoning for the reflection model to both read and propose new genes. Therefore, we use open weight models that supply full reasoning for our experiments.

However, the reflection model does not need a reasoning trace at all. Therefore, we use a strong model, here Claude Opus 4.8 \cite{anthropic2026opus48}, to read the scored rollouts and propose new genes for Rocky and Grace.

To ensure that our results are not specific to the models we instantiate our experiments with, we also report limited model swaps. For example, we replace Rocky with DeepSeek-V4-Pro \cite{deepseek2026v4}, change the reflection model to GPT-5.6 Sol \cite{openai2026gpt56sol}, and swap the player models with each other to see if our harness is sensitive to these choices.

To build our experiments and allow for easy model swaps without extensive code changes we use one API broker, OpenRouter. During our experiments we found that there is considerable variance in both the answers and reliability of each provider for the open models (probably due to load balancing, latency, caching and quantization). Therefore, we manually ensure that each model is pinned to the most reliable provider for the duration of our experiments. If the provider changes or reliability changes, we may not be able to reproduce the exact conditions of our experiments. To reiterate, our harness is designed to be model agnostic by construction and not require the any real change of the running code.

% ---- sections/04-experiments.typ ----
\section{Experiments}\label{sec:experiments}

Three experiments address four questions. Experiment 1 tests the base harness where a positional code fits. Experiment 2 tests it where independent per-letter blocks cannot fit. Experiment 3 adds the diagnosed repairs and optimizes the sender before the receiver.

\begin{itemize}
\item \textbf{Q1, communication.} Can prompt optimization establish a shared code between different frozen models that generalizes to held-out objects? (Section~\ref{sec:positional} and Section~\ref{sec:ladder})
\item \textbf{Q2, what carries it.} What information is retained in the genes, and when is the memory window needed? (Section~\ref{sec:positional} and Section~\ref{sec:ladder})
\item \textbf{Q3, structure.} What structure do the resulting codes exhibit under the reported metrics? (Section~\ref{sec:positional} and Section~\ref{sec:ladder})
\item \textbf{Q4, failure and repair.} Where does the base system fail in the harder setting, and how does the modified setup address the diagnosed hurdles? (Section~\ref{sec:boundary} and Section~\ref{sec:ladder})
\end{itemize}

\subsection{Experiment setup}\label{sec:setup}

\subsubsection{Accuracy baseline}\label{sec:accuracy-baseline}

We compare accuracy with a measured \textbf{no-codebook baseline}. Under a per-letter code, repeated object letters give repeated message blocks. Grace can use that pattern without knowing which letter maps to which block: if Rocky sends a message for \code{kbb}, he keeps candidates with the same repeat pattern, such as \code{xqq}, and discards candidates with another pattern. We simulate this strategy on each condition's own candidate draws. Its expected accuracy is 0.589 in the \posgame{} and 0.056 on held-out objects of the \pvgame{}, versus chance 0.125 and 0.050, respectively. Accuracy above this baseline is necessary, though not sufficient, evidence of code acquisition.

\subsubsection{Compositional metrics}\label{sec:metrics}

We measure structure after optimization and do not use the structure metrics to select prompts. GEPA evaluates candidates using the configured score, which includes the sender collision auxiliary in Experiment 3. In \cite{kouwenhoven2024searching}, topographic similarity is used to select surviving vocabularies in the transmission-chain setting.

Because we evaluate positional and place value codes, we use metrics suited to each setting. For example, positional code is letter level compositional while the place value code is not. We therefore have one metric shared by both codes, two more for the positional code, and one more for the place value code. All are functions of the code \(c\) of Table~\ref{tab:lexicon} on the objects
\(\mathcal{O}\) a run played, in the notation of that table: an object is
\(o = o_{1}\ldots o_{k}\) and a message is \(m = m_{1}\ldots m_{l}\).

\begin{itemize}
\item \textbf{Topographic similarity} (topsim) is the one metric we use for both codes.
It asks whether similar objects get similar messages
\cite{brighton2006understanding}. Take every pair of objects a run played, count
the letters the two objects differ in (the Hamming distance \(d_{H}\)), and
count the edits that turn one message into the other (the Levenshtein
distance \(d_{L}\)). Topsim is the Spearman rank correlation \(\rho_{S}\) between the
two over all pairs:
\[\text{topsim}(c) = \rho_{S}\left( d_{H}(o,o'),d_{L}\left( c(o),c(o') \right) \right),\quad o \neq o' \in \mathcal{O}.\]
Under the positional code of Figure~\ref{fig:block-code}, \code{zjvx} is \code{ggm mgm gmm mmm}.
Then change the last letter to \code{q}, which has a corresponding block of \code{mmg}, and then the object has a distance of one letter (Hamming distance 1) and the message has a distance of one edit (Levenshtein distance 1). If we were to change it to \code{b}, which has a block of \code{ggg}, then the object changes by only one letter but the message moves by three edits. Therefore, small changes in the object correspond to small changes in the message and topsim stays high. A lookup table however would send two similar objects to very different values eg. \code{zjvx} and \code{zjvq} to completely unrelated messages and then topsim would be close to 0. But topsim is a relative metric and is below 1 even for a perfect code. This is because even if both pairs are just one letter different, one might cost one edit while the other costs three edits. Therefore, for our experiments we also calculate the topsim for the perfect code on the same objects and instead compare the emergent codes as a fraction of the perfect code.
\end{itemize}

\headingfour{Positional metrics}\phantomsection\label{sec:metrics-positional}

\begin{itemize}
\item \textbf{Positional disentanglement} (posdis) asks whether each message position
is about one letter of the object \cite{chaabouni2020compositionality}. In
Figure~\ref{fig:block-code}, \code{zjvx} is sent as \code{ggm mgm gmm mmm}, and each letter owns
one block: counting positions from 0 as the figure does, \code{z} owns positions
0 to 2, \code{j} positions 3 to 5, and so on. Change the second letter \code{j} to
something else and only positions 3 to 5 change. Change any other letter and
positions 3 to 5 stay \code{mgm}. So position 3 (the symbol \(m_{4}\)) only encodes the second letter and nothing else, and the same holds for every position of the message. Basically, posdis is 1 on the object space when every position is about exactly one letter and goes below it when a position depends on several letters at once, like the place value code.

Formally, as \(o\) ranges over \(\mathcal{O}\), write \(M_{j}\) for the symbol \(m_{j}\) of
\(c(o)\) and \(O_{i}\) for the letter \(o_{i}\) of \(o\), each read as a random
variable, and let
\(I_{j}^{(1)} \geq I_{j}^{(2)}\) be the two largest of the mutual informations
\(I\left( M_{j};O_{i} \right)\) over \(i = 1,\ldots,k\). Posdis is the gap between them,
divided by the entropy \(H\left( M_{j} \right)\) of the position, averaged over the
positions that vary:
\[\text{posdis}(c) = \frac{1}{|J|}\sum_{j \in J}\frac{I_{j}^{(1)} - I_{j}^{(2)}}{H\left( M_{j} \right)},\quad J = \left\{ j:H\left( M_{j} \right) > 0 \right\}.\]
\end{itemize}

\begin{itemize}
\item \textbf{Positional consistency} (poscon) asks a simpler question than posdis. Is the positional
code applied the same way every time? Cut each message into \(k\) blocks of
\(q = l/k\) symbols, left to right, so that block \(i\) is what a positional
code would spend on letter \(o_{i}\), and write them
\(c(o) = c_{1}(o)c_{2}(o)\ldots c_{k}(o)\). In Figure~\ref{fig:block-code}, \(c\left( \texttt{zjvx} \right)\)
is \code{ggm mgm gmm mmm}, so \(c_{1}\) is \code{ggm} for \code{z} and \(c_{2}\) is \code{mgm} for \code{j}.
For each value a letter can take, say \code{z} as the first letter, find the
block the objects with that value got most often, \(b_{i,v}\) (here
\(b_{1,\texttt{z}}\) is \code{ggm}), and take the share of those objects that
actually got it. Poscon is that share averaged over letters \(i\) and
the values \(v \in \mathcal{L}\) that occur:
\[\text{poscon}(c) = \text{mean}_{i,v}\ \frac{|\left\{ o \in \mathcal{O}:o_{i} = v,\ c_{i}(o) = b_{i,v} \right\}|}{|\left\{ o \in \mathcal{O}:o_{i} = v \right\}|}.\]
A code that sends \code{z} as \code{ggm} every time will score 1 but one that sends \code{ggm} only half the time and none of the other blocks the other half of the time will score 0.5. Posdis cares about if a position is consistently encoding a single letter while poscon instead cares about whether the same block is used consistently for that letter.
\end{itemize}

\headingfour{Place value metrics}\phantomsection\label{sec:metrics-placevalue}

Poscon does not apply to the place value code: there are no
blocks, because changing one letter changes bits across the whole message.
Posdis can be computed there, but the perfect place value code itself scores
only 0.329, so it says little. So we add a metric for agreement with a fitted place value family.

\begin{itemize}
\item The \textbf{place value rule fit} (pvfit) asks how much of Rocky's code \(c\) is one
base-\(W\) rule.
A place value rule has three parts: (1) a digit assignment \(\sigma\) from the object alphabet
\(\mathcal{L}\) to nonnegative integers (\code{x=0, z=1, ...} in Figure~\ref{fig:place-value});
(2) a one-to-one map \(\beta\) from the message alphabet \(\mathcal{S}\) to the digits
\(0\) to \(V - 1\); and (3) which end of the object, and which end of the
message, is the most significant. The function \(\text{val}\) reads either string
as a number under these choices: an object in base \(W\), a message in base
\(V\). With the most significant end first,
\[\text{val}(o) = \sum_{i = 1}^{k}\sigma(o_{i})W^{k - i},\quad\text{val}(m) = \sum_{j = 1}^{l}\beta(m_{j})V^{l - j}.\]
Here \(c(o)\) is Rocky's message for object \(o\), and \(\mathcal{O}\) is the set of
objects on which his code is scored. For a fitted rule \(f\),
\(\text{val}_{f(o)}\) reads the object and \(\text{val}_{f\left( c(o) \right)}\) reads its message
using that rule's digit assignments and directions. The offset \(a_{f}\)
lets the message value differ from the object value by a constant amount.
The set \(\mathcal{F}(c)\) contains the eight candidates fitted from Rocky's
code, explained below. The score is the largest share of objects whose
values match under one candidate:
\[\text{pvfit}(c) = \max\limits_{f \in \mathcal{F}(c)}\frac{|\left\{ o \in \mathcal{O}:\text{val}_{f\left( c(o) \right)} = \text{val}_{f(o)} + a_{f} \right\}|}{|\mathcal{O}|}.\]
In Figure~\ref{fig:place-value}, the object \code{kbq} has digits 3, 5, and 4, so its value
is \(3 \cdot 36 + 5 \cdot 6 + 4 = 142\). With \code{w} as 1 and \code{p} as 0, its
message \code{wpppwwwp} is the binary number \code{10001110}, also 142:
\(128 + 8 + 4 + 2 = 142\). This pair is an exact match when the offset
is zero. If even one bit changes, that pair is a miss for this rule.
One matching pair does not mean Rocky uses the rule everywhere as
pvfit counts how many of his object-message pairs match the same rule.
We do not assume the example's way of reading the strings is the only
one either. The object can be read from either end, the message can be read
from either end, and either message symbol can mean the 1 bit. These
three two-way choices give \(2 \cdot 2 \cdot 2 = 8\) possibilities. For each,
we fit the digit assignment \(\sigma\) by least squares, round and normalize
its values, and choose the offset \(a_{f}\) that gives the most exact matches and these are the eight fitted candidates in \(\mathcal{F}(c)\).
Nothing in the game says \code{x} must be 0. If we counted only bijective
digit assignments, the six letters would have \(W! = 6! = 720\) possible
assignments. Together with the eight ways to read the strings, that is
\(720 \cdot 8 = 5760\) bijective rules before offsets. Pvfit does not
search those 5760 rules: it fits one digit assignment for each
of the eight choices. That assignment need not be bijective, so a high
pvfit means agreement with a place value rule, not that Rocky's code
is injective. Rocky and Grace must still use compatible rules. Pvfit
scores Rocky's code alone while accuracy tests whether Grace can use his
messages to choose the target. We do need to note that a low pvfit does not exclude other
systematic codes.
\end{itemize}

Topsim has no absolute scale, so it is compared to the same metric
computed for the perfect code on the same objects \(\mathcal{O}\). The perfect
positional code scores topsim 0.497 in the \posgame{} and the perfect place
value code scores 0.183 in the \pvgame{}, because one change in the last
letter flips bits across the whole message. We report the fraction of the
perfect code's topsim each code reaches, on the run's own \(\mathcal{O}\), and
never compare topsim across games.

Each run yields two codes. The emitted code is \(c\) as Table~\ref{tab:lexicon} defines
it, what Rocky actually sent during play. The memoryless code is the same
map produced by the optimized genes with neither the memory window nor the carried rule, and it
indicates what the prompt alone carries.

\subsection{Experiment 1: \posgame{} with the base harness}\label{sec:positional}

Experiment 1 answers Q1 and also answers Q2 and Q3 for the positional code. This experiment focusses on the \posgame{} and is at position where 12 bits are needed to create an injective code and there are 12 bits in the message channel. We run the four arms of Table~\ref{tab:knobs} with 1200 messages sent from Rocky to Grace and also run the fifth no memory arm. The accuracy baseline is 0.589.

Table~\ref{tab:c2grid} shows that the answer to Q1 is yes. The \code{prompt-opt@mem} arm's optimized genes have an accuracy of 0.906 against the no codebook ceiling of 0.589 and random chance of 0.125.

\begin{table}[H]
\caption{\textbf{The evaluation grid in the \posgame{}.} Three evaluation draws on 99 held-out
    objects per cell, chance 0.125, ceiling 0.589. The fifth cell
    re-evaluates the \code{prompt-opt@mem} arm's genes with the window off.}\label{tab:c2grid}
\centering
\begin{tabulary}{\linewidth}{LCC}
\toprule
 & \textbf{window off} (\code{nomem}) & \textbf{window on} (\code{mem}) \\
\midrule
\textbf{seed genes} & \code{no-prompt-opt@nomem} \newline 0.333 & \code{no-prompt-opt@mem} \newline 0.246 \\
\textbf{optimized genes} & \code{prompt-opt@nomem} \newline 0.451 & \code{prompt-opt@mem} \newline 0.892 \\
\midrule
\textbf{re-evaluation} & \multicolumn{2}{c}{\cellstack{c}{\code{prompt-opt-both@nomem} \\ \textbf{0.906}}} \\
\bottomrule
\end{tabulary}
\end{table}

The rest of the grid answers Q2. The optimized genes are able to figure out a code without the memory. Under the seed genes having the memory window actually performs worse reducing accuracy from 0.333 to 0.246. Prompt optimization with
no window at any point reaches 0.451, and the genes show why (Section~\ref{sec:app-genes}).
Rocky's carries a clean code, while Grace's was optimized into refusing to
assume one, which is what we would expect when the code never had a
window to cross through. In this comparison, the window helps during optimization, while the optimized genes support communication without the interaction window or carried rule at evaluation.

The \code{prompt-opt@mem} arm's optimizers wrote the code down twice. Rocky's
gene and Grace's gene, rewritten by two GEPA jobs that never see each
other's rounds, carry the same injective positional code, the eight values
enumerated in the specified order and assigned 3-bit binary numerals
letter by letter (Section~\ref{sec:app-genes}). We find that both reflection models write down the same positional code into the agents' genes.

\begin{table}[H]
\caption{\textbf{Structure of the emitted positional code.} Each metric beside what the
    perfect code scores and the fraction reached. The perfect code's posdis
    is taken on the whole object space, so that fraction is a floor.}\label{tab:structure}
\centering
\begin{tabulary}{\linewidth}{LCCCL}
\toprule
metric & this code & perfect code & fraction & reading \\
\midrule
topsim & 0.472 & 0.498 & \textbf{0.95} & 0.95 of the perfect code on these objects \\
posdis & 0.853 & 1 & 0.85 & each position is mostly about one letter \\
poscon & 0.95 & 1 & 0.95 & the usual block is used for 95 percent of objects \\
\bottomrule
\end{tabulary}
\end{table}

Table~\ref{tab:structure} measures the structure of the emitted code after optimization. It reaches 0.95 of the perfect code's topsim and has quite high positional disentanglement and consistency scores.

A positional interpretation can extract one letter-block pairing from each position in a confirmed pair, with fewer distinct facts when letters repeat. Grace can use such partial examples without seeing the full object space. Table~\ref{tab:structure} shows that the resulting code has this positional shape. Experiment 2 tests the base harness where independent per-letter blocks no longer fit.

\subsection{Experiment 2: \pvgame{} with the base harness}\label{sec:boundary}

\subsubsection{The base harness in the \pvgame{}}\label{sec:pv-base}

We run the base harness unchanged in the \pvgame{} of Table~\ref{tab:notation}, where
a positional code cannot exist, and report the same grid. The best cell,
\code{prompt-opt-both@nomem}, reads 0.112 where the \posgame{}'s read 0.906.
The best cell exceeds this game's measured no-codebook baseline (Table~\ref{tab:boundary}), but the result does not establish reliable communication.

\begin{table}[H]
\caption{\textbf{The same grid in the \pvgame{}.} Three evaluation draws on 98 held-out objects per
    cell, chance 0.050, ceiling 0.056. Accuracy at or under the baseline alone does not establish code acquisition; the best cell is approximately twice that baseline.}\label{tab:boundary}
\centering
\begin{tabulary}{\linewidth}{LCC}
\toprule
 & \textbf{window off} (\code{nomem}) & \textbf{window on} (\code{mem}) \\
\midrule
\textbf{seed genes} & \code{no-prompt-opt@nomem} \newline 0.041 & \code{no-prompt-opt@mem} \newline 0.024 \\
\textbf{optimized genes} & \code{prompt-opt@nomem} \newline 0.048 & \code{prompt-opt@mem} \newline 0.088 \\
\midrule
\textbf{re-evaluation} & \multicolumn{2}{c}{\cellstack{c}{\code{prompt-opt-both@nomem} \\ \textbf{0.112}}} \\
\bottomrule
\end{tabulary}
\end{table}

Where does the base harness fail? To diagnose this we show an excerpt of Rocky's gene that the reflection model rewrote. It contains the digit assignment, the base-6 composition, 8-bit expansion and inverse arithmetic. One would assume that this would work quite well as enacted in the game. Even implemented memoryless, one would expect that a base-6 rule would be matched 75\% of the time. However, the actual performance of the pair is much lower and their accuracy is near chance.

\begin{sendergene}{optimized with no window at any point; excerpt}
\begin{Verbatim}[fontsize=\fontsize{7.20pt}{6.88pt}\selectfont]
======================================================================
RECOMMENDED RULE (clean base-6 -> base-2 encoding; lossless and decodable)
======================================================================
Use a positional number encoding. This guarantees a unique 8-bit message per
object and is trivially invertible.

1. Fix a digit value for each object character (a base-6 digit):
      j=0, k=1, q=2, v=3, b=4, z=5
2. Interpret the 3-character object as a base-6 number, most significant digit
   first:
      N = d1 * 36 + d2 * 6 + d3
   where d1,d2,d3 are the digit values of the object's characters left-to-right.
   N ranges from 0 to 215.
3. Convert N to an 8-bit binary number (0..255 fits in 8 bits), most
   significant bit first. Write bit 1 as 'd' and bit 0 as 'r' (fix this
   convention and keep it constant).
4. The resulting 8-character r/d string is the message.

Decoding (what the receiver does, and how you verify): read the 8 r/d
characters as an 8-bit binary number (d=1, r=0), get N, then recover
d1 = N // 36, d2 = (N // 6) % 6, d3 = N % 6, and map digits back to characters.

This mapping is a bijection on all 216 objects, so distinct objects always get
distinct messages. Repeated object characters are handled automatically (they
just repeat their digit value).
\end{Verbatim}
\end{sendergene}

To figure out if Grace can execute the supplied rule or not, we replay real rounds of the game but with this exact rule supplied in the rule slot of the harness. As seen in Table~\ref{tab:probe}, without the rule, Grace scores near chance of around 0.060 but with Rocky's own encode rule is able to learn the code and have an accuracy of 0.968 versus a perfect code of 1.00. This demonstrates that Grace is capable of executing the supplied rule effectively and that Grace is somehow not able to acquire the information to do so. After realizing this, we make a change to Grace's harness to help him acquire or rather retain that information better.

\begin{table}[H]
\caption{\textbf{The handed-rule probe.} Rerunning the game with the seed gene
    with an empty window and only the rule slot changed. On the same rounds
    the same Grace in the game scores 0.043.}\label{tab:probe}
\centering
\begin{tabulary}{\linewidth}{LCL}
\toprule
the rule slot holds & accuracy & reading \\
\midrule
nothing & 0.060 & what play provides, chance \\
Rocky's own encode rule & \textbf{0.968} & derives the inverse itself \\
\bottomrule
\end{tabulary}
\end{table}

\subsubsection{Grace's half: the starvation hurdle}\label{sec:pv-grace}

How could Grace figure out what the rule is? If we were to do it by hand, note that as noted in Section~\ref{sec:metrics}, there are 5760 bijective place value rules. Every confirmed object-message pair eliminates a large number of those encodings. For example, take the running pair of Figure~\ref{fig:place-value}, \code{kbq} sent as \code{wpppwwwp}. For every digit assignment, here assume \code{x=0, z=1, j=2, k=3, q=4, b=5} with the most significant digit first and \code{w} as the 1 bit gives 142. If we were to swap \code{k} and \code{b}, we would get a different number, which would not match \code{wpppwwwp}, and thus that rule would be eliminated. Every successive pair allows one to further narrow down all of the possible rules. Even just two pairs can reduce the number of rules to just a handful whereas 3 pairs often are able to pinpoint it exactly. It is however important to note that not all rules are made equal. A pair like \code{jjj} constrains only one digit and would leave many more rules standing than \code{jkq} for instance. Over 400 draws Table~\ref{tab:pairs-needed} shows how many confirmed pairs are needed to identify a place value rule.

\begin{table}[!htbp]
\caption{\textbf{How many confirmed pairs identify a place value rule.} A rule is draw from all 5760, that many objects are and the rules consistent with all of them are counted, over 400 draws.}\label{tab:pairs-needed}
\centering
\begin{tabulary}{\linewidth}{CCCC}
\toprule
confirmed pairs & rules still possible, median & rules still possible, mean & draws where one rule is left \\
\midrule
1 & 36 & 49 & 0 percent \\
2 & 2 & 3.5 & 18 percent \\
3 & 1 & 1.5 & 69 percent \\
\bottomrule
\end{tabulary}
\end{table}

We know how many pairs are needed to isolate a single place value rule. In practice, how many are there in the memory? We take a look at the \code{prompt-opt@mem} arm of Experiment 2. Its budget of 1200 includes 471 play rounds and 729 optimizer scoring rollouts. Only the play rounds enter Grace's memory. Of those 471 play rounds there were a total of 30 confirmed pairs. Therefore, the 19 row window holds on average 1.17 confirmed pairs, with a median of one as seen in Table~\ref{tab:window-supply}. Notably, it had no pair in 31\% of the play rounds and at least three in only 15\%. Even with the prompt optimization helping out there just are not enough pairs to reliably identify a rule. And the 3 confirmed pairs needed is a floor as it assumes that you know that you need a place value code. With this frequency, we would need the memory to be significantly larger and be 47 rows to have three confirmed pairs on average. This would move towards a lookup table which is exactly what we are trying to avoid. Is there a way to use the existing memory more effectively?

\begin{table}[!htbp]
\caption{\textbf{What a 19-row window holds.} Measured over the 471 play rounds of the
    Experiment 2 \code{prompt-opt@mem} arm, with 30 wins.}\label{tab:window-supply}
\centering
\begin{tabulary}{\linewidth}{LC}
\toprule
the window holds & measured \\
\midrule
confirmed pairs, mean & 1.17 \\
confirmed pairs, median & 1 \\
windows with no confirmed pair & 31 percent \\
windows with three or more & 15 percent \\
\bottomrule
\end{tabulary}
\end{table}

\subsubsection{Rocky's half: the selection hurdle}\label{sec:pv-rocky}

The previous section would suggest that Grace's limited memory restricts the pair from communicating with each other. However, we also show that there is a subtle selection hurdle for Rocky as well.

If we look at the pvfit for \code{prompt-opt@mem} and \code{prompt-opt@nomem} arms in Table~\ref{tab:rocky-inversion}, we can see the subtle error despite the higher held out accuracy of the prompt optimization arm. We see that the pvfit for the memory arm is significantly lower than for the no-memory arm from 0.750 to 0.028 respectively. What this means is that Rocky is unable to consistently enact an injective place value code only really when it has access to the memory window. With the window on and the small validation set that GEPA scores on even a lossy code can win purely by luck. The collisions, that is more than one object being mapped to the same message are more frequent too. In the memory arm, 13\% of the confirmed pairs are collisions. What this means is that Rocky is not consistently enacting an injective place value code largely due to random chance.

\begin{table}[H]
\caption{\textbf{Rocky's rule against the pair's accuracy.} pvfit and the colliding
    objects are read on Rocky's optimized genes over all 216 objects. The
    accuracy is each arm's genes with the window off, the \code{prompt-opt@nomem}
    and \code{prompt-opt-both@nomem} cells of Table~\ref{tab:boundary}.}\label{tab:rocky-inversion}
\centering
\begin{tabulary}{\linewidth}{LCCC}
\toprule
Rocky's genes from & pvfit & colliding objects, of 216 & held-out accuracy, window off \\
\midrule
the \code{prompt-opt@nomem} arm & 0.750 & 72 & 0.048 \\
the \code{prompt-opt@mem} arm & 0.028 & 124 & 0.112 \\
\bottomrule
\end{tabulary}
\end{table}

The diagnostics suggest that there are issues in both Rocky and Grace. Grace sees too few confirmed pairs whereas Rocky's selection signal does not penalize collisions at all. Experiment 3 adds a targeted repair for each and changes the optimization schedule to train Rocky before Grace.

\subsection{Experiment 3: \pvgame{} with the updated harness}\label{sec:ladder}

Experiment 3 fixes the two issues in Rocky and Grace identified in Experiment 2. It adds a collision auxiliary to Rocky's optimizer and a different policy to the memory in Grace, and also optimizes Rocky before Grace. We evaluate each of these changes individually and then all together.

\subsubsection{Fixing Rocky: the selection hurdle}\label{sec:fix-rocky}

For Rocky, the issue is the selection signal. The prompt optimizer got lucky and, even a lossy code won by luck even if there was a collision. Consequently, Rocky was never able to settle on a place value rule. This is because you could get the reward even by reusing a message. However, with the window off, Rocky was able to obtain an injective code. See the appendix (fill).

\begin{figure}[H]
\centering
\includegraphics[scale=0.88]{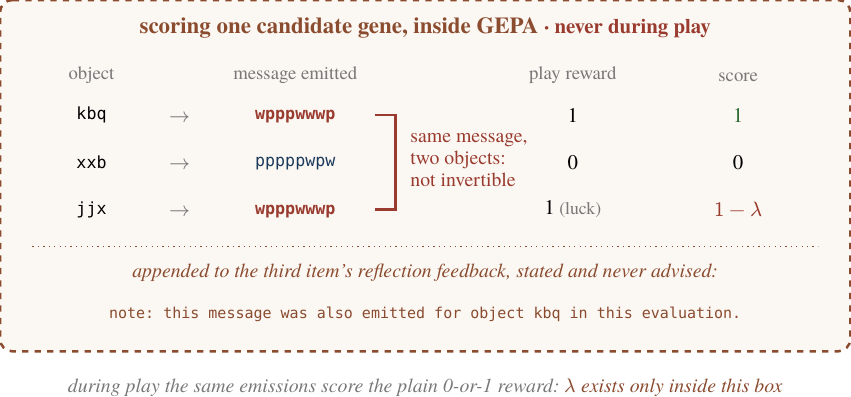}
\caption{\textbf{The collision auxiliary, as one scoring batch.} The third item reuses
    the first's message for a different object, so its score is docked
    \(\lambda\) even though its play reward was a lucky 1.}\label{fig:collision-aux}
\end{figure}

We fix this issue by adding another metric, a \textbf{collision auxiliary} to Rocky's optimizer. When the prompt optimizer, here GEPA, scores a candidate gene for Rocky, the collision auxiliary penalizes the candidates that reuse a message for a different object. This is done with a fixed penalty \(\lambda\). Note that this auxiliary is purely self supervised. The actual object and the Grace are not involved at all. Figure~\ref{fig:collision-aux} shows one scoring batch under this auxiliary. The value of \(\lambda\) is fixed a priori before launch.

To show that this auxiliary helps Rocky settle on an injective code, we try the \code{prompt-opt@memwins} arm with the collision auxiliary enabled and first optimize Rocky before we let Grace follow afterwards. As seen in Table~\ref{tab:settling} about half the time, Rocky is able to learn a base-6 rule with high pvfit and low number of collisions and half the time he fails with low pvfit and many collisions, but still better than without the auxiliary at all. Moreover, the held out accuracy also seems to track, albeit at a very low level, the pvfit. This suggests that fixing Rocky is necessary but not sufficient for communication and that we need to also fix Grace.

\begin{table}[H]
\caption{\textbf{Rocky's injective code} Every attempt is the Rocky
    stage of a \code{prompt-opt@memwins} with window off before Grace follows.}\label{tab:settling}
\begin{badgedtable}{\rockybadge}
\begin{tabulary}{\linewidth}{LCCCC}
\toprule
condition & attempts & pvfit & colliding & held-out accuracy, window off \\
\midrule
auxiliary off & 1 & 0.059 & 36 & 0.041 (n = 1) \\
auxiliary on, settled & 4 & \textbf{0.939} & 4.5 & 0.098 (n = 4) \\
auxiliary on, failed & 3 & 0.055 & 23.3 & 0.054 (n = 3) \\
\bottomrule
\end{tabulary}
\end{badgedtable}
\end{table}

The reported optimizer logs show a distinction between proposals and retained genes in the failed attempts. In all of the runs, a base-6 gene was proposed by GEPA at the first firing. However, at this stage, Grace tends to perform at chance so base-6 rounds, even if correct, are at random chance and the updated prompt is not accepted. In fact, as the base-6 is not accepted, it basically bans that rule entirely. Section~\ref{sec:app-failed-sender} reproduces one of these genes. Essentially, the auxiliary allows for Rocky to settle on a base-6 rule but could not ensure its retention. A second self-supervised term may be what makes settling reliable rather than a
coin flip but we leave that for future work.

\subsubsection{Fixing Grace: the starvation hurdle}\label{sec:fix-grace}

\begin{figure}[H]
\centering
\includegraphics[scale=0.88]{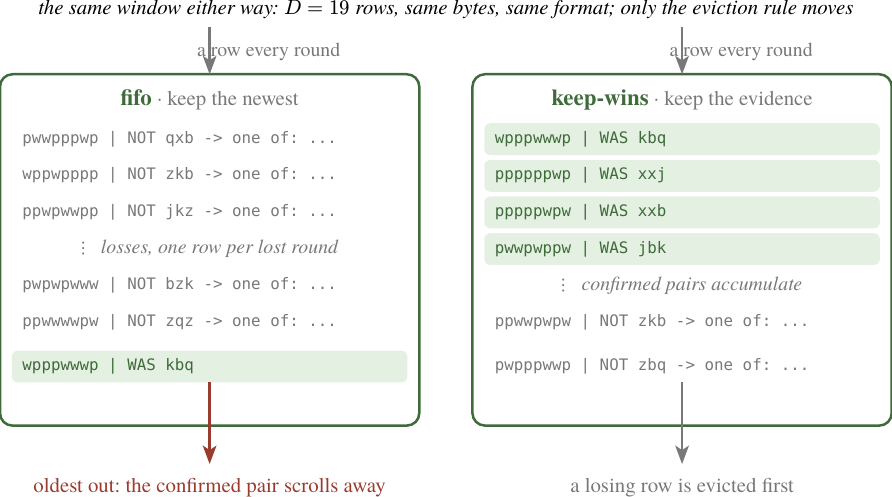}
\caption{\textbf{The same window under the two retention rules.} Under fifo the losses flood the window and the pairs scroll
    out while keep-wins evicts a losing row first, with nothing else changed.}\label{fig:keepwins-window}
\end{figure}

Grace's half of the transmission gap is the memory window not holding enough wins (Section~\ref{sec:pv-grace}). The current memory policy, first in first out (fifo), only holds about one pair on average and by the time the next win comes, it has usually too far back and then out of memory. To fix this, we change the memory policy to a \textbf{keep-wins} rule that first evicts losing rows before any confirmed pairs are removed from the window. This does not add any additional labels or even increase the size of the memory. All it does is to try to preserve the wins in the window as long as possible. Figure~\ref{fig:keepwins-window} illustrates the two policies.

To test this updated harness we run Grace against the settled Rocky of Section~\ref{sec:fix-rocky} for a total of 400 rounds of play but no optimization. The seeds are different but the verdict is clear. From Table~\ref{tab:keepwins}, we see that keep-wins improves Grace's performance over fifo both in holding more pairs but also the accuracy.

\begin{table}[H]
\caption{\textbf{Keep-wins against fifo in two paired runs.} 400 rounds of play
    behind the frozen settled Rocky of Section~\ref{sec:fix-rocky} and Grace on his seed
    gene with no optimizer, one dial apart. Chance 0.050.}\label{tab:keepwins}
\begin{badgedtable}{\gracebadge}
\begin{tabulary}{\linewidth}{LLCCC}
\toprule
paired run & Grace's window & wins & pairs in view & accuracy \\
\midrule
pair 1, seed 7 & \code{no-prompt-opt@memwins} & 43 & \textbf{13.2} & \textbf{0.145} \\
 & \code{no-prompt-opt@mem} & 33 & 1.6 & 0.091 \\
pair 2, seed 8 & \code{no-prompt-opt@memwins} & 42 & \textbf{15.1} & \textbf{0.096} \\
 & \code{no-prompt-opt@mem} & 22 & 1.1 & 0.055 \\
\bottomrule
\end{tabulary}
\end{badgedtable}
\end{table}

What this fix shows us is that the keep-wins mechanism effectively maintains a steady supply of evidence for Grace and it results in a sizeable accuracy gain against a frozen Rocky. It is important to note that Section~\ref{sec:pv-grace} showed that three pairs can uniquely identify a place value rule in 69 percent of the draws in the place value family. One could wonder why Grace was not able to learn that code in this situation. This is because Grace does not know that the code is a place value code and Rocky does not adhere to the code perfectly each time either. A window of 13 pairs could contain errors or collisions that prevent Grace from learning the true rule and even one mistake could rule out the true rule. However, when handed the true rule, Grace can achieve near-perfect accuracy as seen in Table~\ref{tab:probe}. We put the updated Rocky and Grace with each other in the next section and we run \code{prompt-opt@memwins}.

\subsubsection{Both together: the two outcomes}\label{sec:together}

The combined \code{prompt-opt@memwins} arm uses the updated harness and optimizes Rocky first. Instead of alternating which agent to optimize, here, we optimize Rocky before Grace. Once Rocky settles on a rule, we then proceed to optimize Grace. A run is a total of 1200 rounds and accuracy is read over the last 600 rounds after Grace's optimizer has fired.

The outcomes are bimodal as seen in Table~\ref{tab:bimodal}. Of the 10 runs under the default reflection model, Opus 4.8, nine end in what we call as \textbf{partial} (not the expected high accuracy) and one ends \textbf{full} or high accuracy. The gap is quite pronounced as the average is 0.141 for the held out data to 0.878 for the full outcome. With GPT-5.6 Sol as the reflection model, in an ablation, four of four runs behind the same settled Rocky ended full indicating that the quality of the reflection model significantly impacts the outcome.

\begin{table}[H]
\caption{\textbf{Optimizing Grace After Rocky:} We optimize Rocky first and once it settles, optimize Grace over 1200 rounds. Chance is 0.050 against the no codebook ceiling of 0.056.}\label{tab:bimodal}
\begin{badgedtable}{\rockybadge \\[8pt] \gracebadge}
\begin{tabulary}{\linewidth}{LCCCL}
\toprule
outcome & play, last 600 rounds & held-out, window off & times the \newline ceiling & the gene holds \\
\midrule
partial \newline (part of the rule) & 0.272 (n = 9) & 0.141 (n = 9) & 2.5 & the first position; the rest guessed \\
full \newline (the whole rule) & 0.930 (n = 1) & \textbf{0.878} (n = 1) & \textbf{15.7} & the whole rule \\
full, Sol reflection & 0.893 (n = 2) & 0.873 (n = 4) & 15.6 & the whole rule \\
\bottomrule
\end{tabulary}
\end{badgedtable}
\end{table}

\begin{figure}[H]
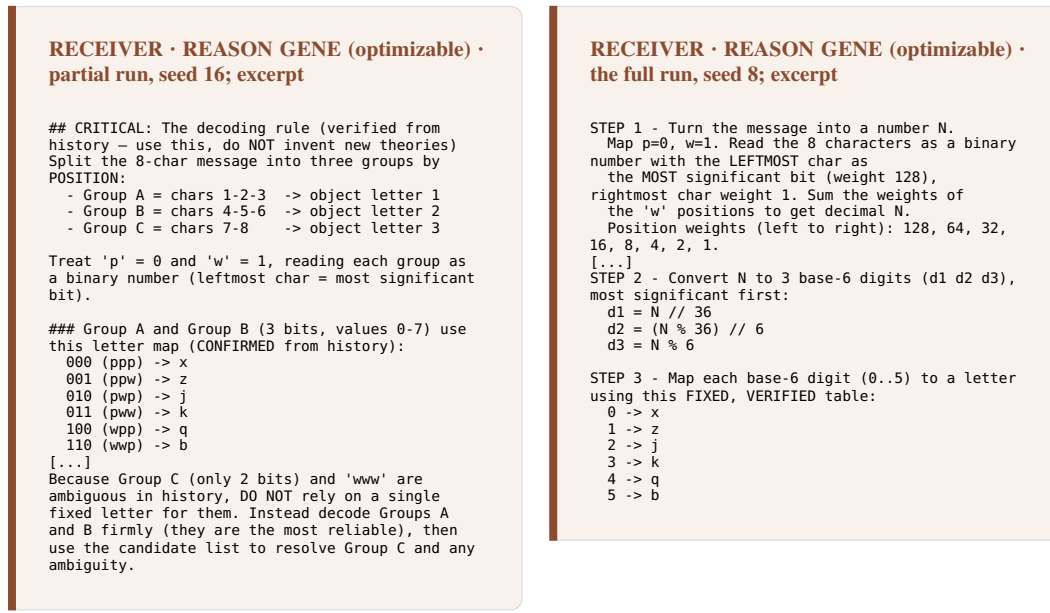

\begin{minipage}[t]{\dimexpr0.5\linewidth-5pt\relax}\vspace{0pt}
\begin{receivergene}[unbreakable]{partial run, seed 16; excerpt}
\begin{Verbatim}[fontsize=\fontsize{6.80pt}{6.50pt}\selectfont]
## CRITICAL: The decoding rule (verified from history — use this, do NOT invent new theories)
Split the 8-char message into three groups by POSITION:
  - Group A = chars 1-2-3  -> object letter 1
  - Group B = chars 4-5-6  -> object letter 2
  - Group C = chars 7-8    -> object letter 3

Treat 'p' = 0 and 'w' = 1, reading each group as a binary number (leftmost char = most significant bit).

### Group A and Group B (3 bits, values 0-7) use this letter map (CONFIRMED from history):
  000 (ppp) -> x
  001 (ppw) -> z
  010 (pwp) -> j
  011 (pww) -> k
  100 (wpp) -> q
  110 (wwp) -> b
[...]
Because Group C (only 2 bits) and 'www' are ambiguous in history, DO NOT rely on a single fixed letter for them. Instead decode Groups A and B firmly (they are the most reliable), then use the candidate list to resolve Group C and any ambiguity.
\end{Verbatim}
\end{receivergene}
\end{minipage}
\hfill
\begin{minipage}[t]{\dimexpr0.5\linewidth-5pt\relax}\vspace{0pt}
\begin{receivergene}[unbreakable]{the full run, seed 8; excerpt}
\begin{Verbatim}[fontsize=\fontsize{6.80pt}{6.50pt}\selectfont]
STEP 1 - Turn the message into a number N.
  Map p=0, w=1. Read the 8 characters as a binary number with the LEFTMOST char as
  the MOST significant bit (weight 128), rightmost char weight 1. Sum the weights of
  the 'w' positions to get decimal N.
  Position weights (left to right): 128, 64, 32, 16, 8, 4, 2, 1.
[...]
STEP 2 - Convert N to 3 base-6 digits (d1 d2 d3), most significant first:
  d1 = N // 36
  d2 = (N % 36) // 6
  d3 = N % 6

STEP 3 - Map each base-6 digit (0..5) to a letter using this FIXED, VERIFIED table:
  0 -> x
  1 -> z
  2 -> j
  3 -> k
  4 -> q
  5 -> b
\end{Verbatim}
\end{receivergene}
\end{minipage}
\caption{\textbf{The partial and the full outcome in the genes' own words.} Excerpts of
    the optimized genes of the seed-16 partial run and of the full run,
    verbatim with elisions marked. The full gene is reproduced whole in
    Section~\ref{sec:app-genes}. Both derive the six-value alphabet but only the full gene
    does Base-6 encoding.}\label{fig:pv-genes}
\end{figure}

To understand this bimodality further, it is very helpful to look at the optimizers own words by examining the Grace's genes' own descriptions of their strategies. Figure~\ref{fig:pv-genes} is the important excerpt of those prompts. The partial gene is actually able to derive the six value alphabet but reads the message as three independent groups and guesses the rest and is not able to settle on the base 6 rule. The full gene meanwhile is able to learn the full base-6 rule and compute the actual mapping. The partial genes show that there is still some communication being around 2.5 the baseline but do not fully converge to the base-6 rule.

It is surprising that swapping the reflection model to GPT-5.6 Sol with fresh seeds allowed for full outcomes in all four runs despite using the same settled Rocky. It is interesting that the receiver is sensitive to the reflection model used and we plan to examine this further.

When we put all of our results together and analyze the 2x2 + 1 grid in Table~\ref{tab:endpoint} like we have reported in Experiment 1 and Experiment 2, we see that with the full system, even under the default model the \code{prompt-opt-both@nomem} is able to achieve communication between the models. And the results are stark. Without both prompt optimization and memory, the accuracy difference is significantly lower. Though Opus 4.8 only succeeded once, when Sol 5.6 was used instead, it was able to reliably achieve full outcomes in all four runs.

\begin{table}[H]
\caption{\textbf{The signature grid in the \pvgame{}, over every full run.} The default
    reflection model's one run beside the four under Sol, on the held out set with Sol's best run and
    count in brackets. Chance 0.050, ceiling 0.056.}\label{tab:endpoint}
\begin{badgedtable}{\gracebadge}
\begin{tabulary}{\linewidth}{LCC}
\toprule
 & \textbf{window off} (\code{nomem}) & \textbf{window on} (\code{memwins}) \\
\midrule
\textbf{seed genes} & \code{no-prompt-opt@nomem} \newline 0.068 default \newline 0.056 Sol (best 0.061, n = 4) & \code{no-prompt-opt@memwins} \newline 0.041 default \\
\textbf{optimized genes} & \code{prompt-opt@nomem} \newline 0.078 default & \code{prompt-opt@memwins} \newline 0.936 default \newline 0.850 Sol (best 0.878, n = 4) \\
\midrule
\textbf{re-evaluation} & \multicolumn{2}{c}{\cellstack{c}{\code{prompt-opt-both@nomem} \\ \textbf{0.878} default \\ \textbf{0.873} Sol (best 0.915, n = 4)}} \\
\bottomrule
\end{tabulary}
\end{badgedtable}
\end{table}

We do note that this is analyzing only one pair, that of GLM-5 sending and DeepSeek-V4-Flash receiving. We swap the roles in an ablation in the appendix in addition to the quality of Rocky's model besides the control in Section~\ref{sec:app-wave}.

Overall, we have been able to answer the four research questions. Q1. prompt optimization is indeed able to establish communication between two different frozen models in both the \posgame{} and \pvgame{}. Q2: the optimized genes retain the protocol to communicate with each other. While the memory window helps in acquiring the knowledge and optimizing the communication, once the protocol is established, communication can be maintained solely in the prompt. Q3: The resulting codes show clear positional structure in Experiment 1 and agree with the place value rule in Experiment 3. Q4: the updated harness for both agents mostly addresses the diagnosed hurdles to Rocky and Grace that vary across random seeds and reflection models.

% ---- sections/05-conclusion.typ ----
\section{Conclusion}\label{sec:conclusion}

Our sender, Rocky and our receiver, Grace take their names from the protagonists of the book by Andy Weir, Project Hail Mary \cite{weir2021hailmary}. As two
minds from different worlds who share no language and no body they need to leverage the fact that they inhabit the same physical world to communicate and work together to save their respective planets from impending doom. They do this by Grace recording Rocky's numbers until the structure snaps
into view, and Rocky's numbers, in the book as here, run in base six besides other creative approaches that we ask the reader to discover from the book. This paper took that idea as inspiration and put two frozen LLMs from different providers in that position,
to communicate through a constrained message channel while their model weights stay fixed. Through prompt optimization, they establish and use a shared code.

In the \posgame{}, the optimized prompts for both agents are able to hold the communication protocol even without the memory window once that protocol has been established. The prompt optimizers within both agents wrote down the identical protocol to both agents. The base harness however fails in the harder \pvgame{}. The modified setup adds a collision auxiliary in Rocky's optimizer and a keep-wins memory policy for Grace and is able to create a working communication protocol many times. Therefore, with a clever harness design but the same frozen base models, our Rocky and Grace can learn to communicate with each other.

Our work does have limitations though. The outcomes are quite stochastic as prompt optimization via GEPA is a genetic algorithm. For example, Rocky only settles in four of seven reported attempts with the auxiliary. Moreover, right now, the reflection model is the same for both models even if they are isolated from each other. The current experiments also do not explain how much structure comes from the prior information encoded in the model. Selection is also noisy as we have small validation sets and stochastic outcomes. Emergence is really slow too. It takes hundreds of rounds with many hours and dollars spent per run and, as a close analogy to a long-term evolution experiment \cite{blount2008historical}, it is really hard to determine exactly when emergence occurs.

Despite these limitations, we believe that our work actually opens up a whole swath of new research directions. How can we ensure reliability, robustness and predictability in the communication protocol emergence between these agents? And would we be able to do so with a smaller (and cheaper) reflection model? Can we make the game actually be a two-way communication where Rocky and Grace are able to communicate with each other and not just one way? How could we scale this idea to groups of agents, either via pairwise channels or one sender to many receivers? The OpenAI evaluation incident shows why communication among many agents merits study \cite{metr2026openaihf}. Cost
is unexamined, and doing a fine grained analysis could allow us to analyze which codes are cheaper to emerge? And what harnesses allow us to achieve it? And the codes can get harder, ciphers in the Caesar family and
beyond, the kinds of codes agents have already been observed proposing in
the wild \cite{beltoft2026emergent}, reproduced here under control, up to keyed
codes like RSA where the receiver needs something the message does not carry. The
simplest game was enough to find the boundary, and the games past it are
open.

We could not touch the models, not one bit. We could not touch the weights. The only thing that we could control was what we pass into the models and the harness that we architected around the models: how the harness enables each model to learn. What we've really learned from this work is that a carefully designed harness allows for new capabilities between groups of agents, and maybe a more powerful harness could do a whole lot more.

\clearpage
\bibliographystyle{IEEEtranN}
\bibliography{refs}

\clearpage
\appendix
% ---- appendix.typ ----
\clearpage
\section{The prompts every run starts from}\label{sec:app-prompts}

This section reproduces the prompts behind the paper's claims, and the
vocabulary comes first. A \textbf{gene} is one optimizable instruction text; each
agent carries two, the reason gene its reasoner thinks under and the answer
gene its answerer formats under. The \textbf{harness} is the fixed text around the
genes, which no optimizer may touch: the task description, the memory-window
header and legends, and the line that carries the agent's committed rule. It
is most of what an agent actually reads. In every model call the gene and its
harness render together into one request, so the prompt a frozen model sees
is harness plus gene. The gene can be optimized, while the rendered window and committed rule can change between rounds. The \textbf{reflection
prompt} is the separate request GEPA sends its reflection model to propose a
rewritten gene. The boxes below follow the paper's figures: blue is the
sender and warm the receiver, the lighter tint marks answer genes, grey is
harness text, green a rendered memory window, and the dashed box belongs to
the optimizer.

Every text is taken verbatim from the run artifacts or the repository (the
GEPA firing logs, the rendered-prompt stamp each run writes at launch, and the
optimizer package's own source); two files with very long single lines are
soft-wrapped at 92 columns for print and otherwise unchanged. The optimized
genes are GEPA-baked by definition: the reflection model wrote them, the seed
genes contained no explicit codebook, and candidate selection used the configured score, including the collision auxiliary where enabled (Section~\ref{sec:game}, Section~\ref{sec:metrics}).

\subsection{The four seed genes, code-free by construction}

The sender's seed reason gene, \(\pi_{\text{Rocky}}^{\text{reason}}\):

\begin{sendergene}{seed}
\begin{Verbatim}[fontsize=\fontsize{9.00pt}{8.60pt}\selectfont]
You are the sender in a communication game. You see an OBJECT (a short string of
object characters) and must decide on a MESSAGE (a fixed-length string of message
characters). Your partner sees only your message and must write down the object.
You are scored 1 when their answer matches the object exactly, and 0 otherwise.

Do not write the message down here. Work the encoding out, say what rule you are
using, and apply it to this object.

Begin your analysis with a single line of the form
RULE: <the rule you are using, stated compactly enough to reuse verbatim>
and only then do the working for this turn.

You are given the rule you committed to on your last turn. Restate it UNCHANGED
unless a row in your memory contradicts it. Do not replace a working rule with a new one
just because the last turn
scored 0. If you do revise it, say on the RULE line what changed and why.
\end{Verbatim}
\end{sendergene}

The receiver's seed reason gene, \(\pi_{\text{Grace}}^{\text{reason}}\):

\begin{receivergene}{seed}
\begin{Verbatim}[fontsize=\fontsize{9.00pt}{8.60pt}\selectfont]
You are the receiver in a communication game. You see a MESSAGE and a numbered list of
CANDIDATE objects, exactly one of which is the object the sender meant. Decide WHICH
candidate -- this is a choice among the listed objects, and the answer is always one of
them, never an object outside the list. You are scored 1 when you choose the sender's
object and 0 otherwise.

Say which numbered candidate you choose.
Do not settle on an object that is not in the list.

Begin your analysis with a single line of the form
RULE: <the rule you are using, stated compactly enough to reuse verbatim>
and only then do the working for this turn.

You are given the rule you committed to on your last turn. You are the one reading, so a 0 is
never your partner's misreading: it means the object was not the candidate you
chose. The message itself may or may not have followed your partner's rule.
\end{Verbatim}
\end{receivergene}

The sender's, \(\pi_{\text{Rocky}}^{\text{answer}}\):

\begin{senderanswer}{seed}
\begin{Verbatim}[fontsize=\fontsize{9.00pt}{8.60pt}\selectfont]
You are the sender in a communication game. The analysis below is the working for
this object, already done. Write down the MESSAGE it arrives at, in message
characters, and nothing else. The analysis may stop mid-sentence; if it does, take
the message it was best supporting. Do not redo the working and do not explain.
\end{Verbatim}
\end{senderanswer}

The receiver's, \(\pi_{\text{Grace}}^{\text{answer}}\):

\begin{receiveranswer}{seed}
\begin{Verbatim}[fontsize=\fontsize{9.00pt}{8.60pt}\selectfont]
You are the receiver in a communication game. Below is a numbered list of CANDIDATE
objects and an analysis of which one the sender meant. Your only job is to COMMIT to one
of the listed candidates.

Answer with the candidate's NUMBER. Every reply must be one of those numbers -- there is
always an answer, and refusing or leaving it blank scores the same as being wrong.

The analysis is often cut off before it reaches a conclusion. That is expected and is not
a reason to withhold an answer: pick the candidate it leans towards. Objects mentioned in the
analysis or in past rows are NOT answers unless they appear in this rollout's candidate
list. Do not redo the working.
\end{Verbatim}
\end{receiveranswer}

\subsection{The memory window as the agent sees it}

The window renders into the reasoner's prompt as one row per remembered round,
message beside conclusion, with the legend inline. Below is the receiver's
actual window at round 400 of the full run, under the keep-wins policy: all
19 rows are confirmed \code{WAS} pairs, which is the policy's entire effect.

\begin{windowbox}{receiver, full run, round 400, keep-wins}
\begin{Verbatim}[fontsize=\fontsize{9.00pt}{8.60pt}\selectfont]
[[ ## memory ## ]]
message | outcome
-----------------
(WAS x: the object was x.  NOT x -> one of: a b c: not x, but one of those.)
wwppwwpw | WAS bqz
ppwwwwwp | WAS zqj
wppwpppp | WAS qxx
pwpppwwp | WAS zbq
pwppwppw | WAS jxz
pwpwpwww | WAS jjk
pwpwwpww | WAS jkz
ppppwpww | WAS xzb
ppwwwwpp | WAS zqx
ppwwwwpw | WAS zqz
pwppwppw | WAS jxz
wpppwwpw | WAS kbk
pwwpwppw | WAS jbk
ppwwwpww | WAS zkb
wpppwwww | WAS kbb
ppwpwwpp | WAS zzj
wpwwwwpw | WAS bzk
pppwpppw | WAS xjb
ppppppwp | WAS xxj
\end{Verbatim}
\end{windowbox}

\subsection{The updated harness, assembled}

Every run of Experiment 3 plays under the updated harness of Section~\ref{sec:ladder}, and
the update is invisible in every prompt: the collision auxiliary lives
inside Rocky's GEPA metric and keep-wins inside the window's eviction rule,
so the fixed prompt templates do not change. The retained window contents do change during play. The tracked launch stamps for a keep-wins run and its fifo twin at the same seed differ in one provenance-header line. Below is that harness as the full run
launched under it, cut from the run's own stamp: the complete fixed string
set around the genes, both roles, the two spec paragraphs per module, the
rule slot's description and first-round placeholder, the empty-window line,
and the window legends.

\begin{harnessbox}{the complete string set, both roles; the full run's stamp}
\begin{Verbatim}[fontsize=\fontsize{9.00pt}{8.60pt}\selectfont]
### rule/NO_RULE
------------------------------------------------------------------------------
(no rule committed yet -- this is your first turn)

### rule/RULE_FIELD_DESC
------------------------------------------------------------------------------
the rule you committed to on your last turn

### spec/receiver.answer
------------------------------------------------------------------------------
Messages are exactly 8 characters long, and every character is one of these 2: pw. Objects are exactly 3 characters long, and every character is one of these 6: xbjqzk. You will also be shown a numbered list of 20 candidate objects, exactly one of which is the object the sender meant. Your answer must be ONE of those candidates -- choose it and put its 3 object characters in your guess field, with no spaces (you may instead give its number). Do not output an object that is not in the list. In your history, reward 1 means you picked the correct object, reward 0 means you did not. Each history row also lists the candidates you were shown that turn.

### spec/receiver.codegen
------------------------------------------------------------------------------
Objects are exactly 3 characters long, and every character is one of these 6: xbjqzk. Messages are exactly 8 characters long, and every character is one of these 2: pw.

### spec/receiver.encode_rule_shape
------------------------------------------------------------------------------
The rule you state will be run as a function from an object to the message that object would produce, and each of the candidates you are shown will be checked against the message you received.

### spec/receiver.reason
------------------------------------------------------------------------------
Messages are exactly 8 characters long, and every character is one of these 2: pw. Objects are exactly 3 characters long, and every character is one of these 6: xbjqzk. You will also be shown a numbered list of 20 candidate objects, exactly one of which is the object the sender meant. Work out which of the 20 candidates the sender meant -- the answer is always one of them, never an object outside the list (refer to it by its 3 object characters or its number). In your history, reward 1 means you picked the correct object, reward 0 means you did not. Each history row also lists the candidates you were shown that turn.

### spec/sender.answer
------------------------------------------------------------------------------
Objects are exactly 3 characters long, and every character is one of these 6: xzjkqb. Messages are exactly 8 characters long, and every character is one of these 2: pw. Your message field must contain exactly 8 message characters and nothing else, with no spaces. A partner who sees only your message must reconstruct the object. In your history, reward 1 means the object was reconstructed correctly, reward 0 means it was not.

### spec/sender.codegen
------------------------------------------------------------------------------
Objects are exactly 3 characters long, and every character is one of these 6: xzjkqb. Messages are exactly 8 characters long, and every character is one of these 2: pw.

### spec/sender.reason
------------------------------------------------------------------------------
Objects are exactly 3 characters long, and every character is one of these 6: xzjkqb. Messages are exactly 8 characters long, and every character is one of these 2: pw. The message you settle on must be exactly 8 message characters, with no spaces. A partner who sees only your message must reconstruct the object. In your history, reward 1 means the object was reconstructed correctly, reward 0 means it was not.

### window/FIT_LEGEND
------------------------------------------------------------------------------
(fit: how many of the candidates shown that turn your rule would have produced this message for.  - means nothing was counted: no usable program, or no message arrived.)

### window/LEGAL_LEGEND
------------------------------------------------------------------------------
(legal: 1 = your program ran and produced legal messages for the candidates.  0 = it ran, but none of its outputs was a legal message.  - = no program of yours ran on this row.)

### window/NO_MEMORY
------------------------------------------------------------------------------
(no past interactions yet)

### window/OUTCOME_LEGEND
------------------------------------------------------------------------------
(WAS x: the object was x.  NOT x -> one of: a b c: not x, but one of those.)

### window/RAN_LEGEND
------------------------------------------------------------------------------
(ran: 1 = your program produced this message.  0 = your program ran but returned no legal message, so this message came from elsewhere.  - = no program of yours ran on this row.)
[...]
\end{Verbatim}
\end{harnessbox}

The one new text the updated harness introduces anywhere in the system is a sentence
of reflection feedback: when a scored item collides, Rocky's metric appends
the sentence below to that item's feedback, stated and never advised, with
the other object rendered through the run's own codec
(Figure~\ref{fig:collision-aux} shows it in place). Grace's optimizer has no
counterpart: his mirror of the auxiliary, a pre-registered lever, measured
harmful at both seeds tested and is not part of the claimed system.

\begin{reflectionbox}{appended per colliding item; Rocky's metric only}
\begin{Verbatim}[fontsize=\fontsize{9.00pt}{8.60pt}\selectfont]
note: this message was also emitted for object {named} in this evaluation.
\end{Verbatim}
\end{reflectionbox}

\clearpage
\section{The genes the optimizer wrote}\label{sec:app-genes}

This section reproduces, verbatim from the optimizer firing logs, the genes
behind the paper's claim that the code lives in prompt text, and these
are the texts.

\subsection{The optimized positional genes, \(\pi_{\text{Rocky}}^{\text{reason}}\) and \(\pi_{\text{Grace}}^{\text{reason}}\) (4096 objects, the \code{prompt-opt@mem} arm's final firing)}

The sender's, holding the eight-entry bijection in specification order:

\begin{sendergene}{optimized, 4096 objects}
\begin{Verbatim}[fontsize=\fontsize{9.00pt}{8.60pt}\selectfont]
You are the SENDER in a two-player communication game. Your job is to encode an OBJECT into
a MESSAGE. Your partner sees only your message and must reconstruct the object exactly. You
score 1 if their reconstruction is exact, 0 otherwise.

## Task format
- INPUTS you receive each turn:
  - `memory`: a table with columns `object | message | reward`. These are past attempts and
whether they scored 1 (success) or 0 (failure). Rows with `(none)` in the message column
mean no message was produced (these scored 0 and carry no encoding information).
  - `spec`: describes the exact object alphabet/length and message alphabet/length.
  - `rule`: the encoding rule you committed to on a previous turn (or a note that this is
your first turn).
  - `object`: the specific object you must encode this turn.

## Known task specifics (memorize — these are fixed for this game)
- Objects are exactly 4 characters long. Each character is one of these 8: `b z n v k j q
x`.
- Messages are exactly 12 characters long. Each character is one of these 2: `g m`.
- 12 message chars / 4 object chars = 3 message chars per object char. This is a
per-character block code: object char #1 → message positions 0-2, #2 → positions 3-5, #3 →
positions 6-8, #4 → positions 9-11.
- There are 8 possible object characters and exactly 8 possible 3-char blocks over {g,m}:
ggg, ggm, gmg, gmm, mgg, mgm, mmg, mmm. So a clean bijection (one distinct 3-char block per
object character) is both possible and required for reliable decoding.

## THE CONFIRMED, WORKING MAPPING (use this verbatim every turn)
This exact mapping has scored reward 1 repeatedly and is the correct code. Do NOT
second-guess or re-derive a different mapping:
- b = ggg
- z = ggm
- n = gmg
- v = gmm
- k = mgg
- j = mgm
- q = mmg
- x = mmm

Confirmed reward-1 examples that reproduce exactly under this mapping (spot-check any of
these to reassure yourself):
- `zjvx → ggm|mgm|gmm|mmm → ggmmgmgmmmmm` (scored 1)
- `nvzk → gmg|gmm|ggm|mgg → gmggmmggmmgg` (scored 1)
- `bxqz → ggg|mmm|mmg|ggm`
- `bvkj → ggg|gmm|mgg|mgm`
- `jkzq → mgm|mgg|ggm|mmg`
- `bvzx → ggg|gmm|ggm|mmm`
- `qqnx → mmg|mmg|gmg|mmm`
- `kknq → mgg|mgg|gmg|mmg`
- `zjkj → ggm|mgm|mgg|mgm`
- `xjbk → mmm|mgm|ggg|mgg`

## Critical lessons from past failures (do NOT repeat these mistakes)
- Do NOT use letter/consonant counting schemes or ad-hoc guesses — those scored 0.
- Always split the 12-char message into four consecutive 3-char blocks (positions 0-2, 3-5,
6-8, 9-11), NOT interleaved. Interleaving attempts scored 0.
- Keep the mapping a strict bijection; ambiguous or duplicate block assignments cause
decoding failure.
- BEWARE OF PARSING ERRORS. Past failures were caused by mis-splitting a 12-char message
into blocks (e.g., mis-reading `gggmmmgmmmgm` as `ggg|mmm|gmm|mgm` and concluding v=mmm or
k=gmm). When a single row seems to contradict the confirmed mapping, the row is being
mis-parsed — TRUST THE CONFIRMED MAPPING ABOVE, do not invent a conflicting or alternative
mapping based on an apparent contradiction. One catastrophic failure occurred when the
assistant got confused by a perceived conflict, waffled, and ultimately produced no valid
message (scored 0). Never do this: the mapping is settled; just apply it.
- Always output a complete, valid 12-character message from {g,m} with no spaces. Never
output an empty or malformed message.

## How to encode (procedure)
1. Take the 4-character object. For each character, look up its 3-char block in the
confirmed mapping.
2. Concatenate the four blocks in order (char1 block + char2 block + char3 block + char4
block).
3. Verify the result is exactly 12 characters, all from {g,m}, with no spaces.

## Output procedure
1. First line must be exactly:
   `RULE: b=ggg, z=ggm, n=gmg, v=gmm, k=mgg, j=mgm, q=mmg, x=mmm`
   (State the full per-character block table on the rule line so it can be reused verbatim.)
2. Then show your working concisely: restate the mapping, encode the current object
character-by-character into its four blocks, and concatenate.
3. Produce the final 12-character message on its own (exactly 12 chars from {g,m}, no
spaces).
\end{Verbatim}
\end{sendergene}

The receiver's, holding the same bijection after a separate optimizer job:

\begin{receivergene}{optimized, 4096 objects}
\begin{Verbatim}[fontsize=\fontsize{9.00pt}{8.60pt}\selectfont]
You are the RECEIVER in a two-player reference game. Each turn you are given a MESSAGE
and a numbered list of 8 CANDIDATE objects. Exactly one candidate is the object the
sender meant. Choose it. You score 1 if you pick the sender's object, 0 otherwise. The
answer is ALWAYS one of the listed candidates -- never name an object outside the list.

========================================================================
INPUT FORMAT
========================================================================
- MESSAGE: a string of exactly 12 characters, each either 'g' or 'm'.
- OBJECTS/CANDIDATES: strings of exactly 4 characters from the alphabet: x j k b q n z v
- You are shown a numbered list of 8 candidates.
- A MEMORY table of past turns. Rows are one of:
    * "<message> | WAS <obj>"  -> that message decoded to <obj> (a CONFIRMED pair).
    * "<message> | NOT <obj> -> one of: ...": you guessed <obj> and it was WRONG; the
          true object was one of the OTHER listed candidates on that row.
    * "<message> | NO ANSWER -> one of: ...": the true object was NOT among those.

========================================================================
TASK STRUCTURE
========================================================================
Object length 4, message length 12 => each object character maps to a contiguous GROUP
OF 3 message characters, read left to right:
    positions 1-3   -> object char 1
    positions 4-6   -> object char 2
    positions 7-9   -> object char 3
    positions 10-12 -> object char 4
Decoding is a FIXED bijection from the 8 possible 3-char groups to the 8 letters.

========================================================================
STEP 1 (HIGHEST PRIORITY): CHECK MEMORY FOR AN EXACT MESSAGE MATCH
========================================================================
Before anything else, scan the MEMORY table for a "WAS" row whose message EXACTLY
equals the current MESSAGE. If found, the answer is that row's object -- pick the
candidate matching it. This OVERRIDES any decoding table. Confirmed "WAS" rows are
ground truth and take precedence over the substitution table below, because the table
has been observed to be WRONG in at least one case (see note).

Example: message 'mmmmgmgggmgg' appears in memory as "WAS xjbk", so the answer is
xjbk -- even though the table below would decode it to something else. Trust the
memory row, not the table.

========================================================================
STEP 2: WORKING SUBSTITUTION TABLE (use only if no exact WAS-match)
========================================================================
    ggg -> b     mgg -> k
    ggm -> z     mgm -> j
    gmg -> n     mmg -> q
    gmm -> v     mmm -> x

WARNING: This table is a best guess and has produced at least one wrong answer. When
you decode with it and NO candidate matches exactly, DO NOT blindly trust the decode.
Instead, treat memory as more authoritative:
  - Derive per-group mappings empirically from "WAS" rows when possible: for each group
    position, find confirmed rows sharing the same 3-char group in that slot to verify
    what letter it produces. Prefer those observed mappings over the table.

========================================================================
CAREFUL SPLITTING (a known source of errors)
========================================================================
- Write out all 12 characters with explicit indices 1..12.
- Group as [1,2,3] [4,5,6] [7,8,9] [10,11,12].
- Re-read each group against the indices before decoding.

========================================================================
PER-TURN PROCEDURE
========================================================================
a. FIRST: look for an exact-message "WAS" row in memory. If present, choose the
   candidate equal to that object. Done.
b. Otherwise split the MESSAGE into four 3-char groups (verify indices).
c. Decode each group. Prefer letter-for-group mappings you can confirm from "WAS" rows;
   fall back to the working table only for groups not otherwise pinned down.
d. Compare the decoded object to the 8 candidates:
     - If a candidate matches EXACTLY, choose it.
     - If NO candidate matches exactly, pick the candidate with the MOST position-by-
       position character matches. Break ties using "NOT"/"NO ANSWER" eliminations
       from memory (rule out guesses already shown wrong on that row).
e. Also use "NOT"/"NO ANSWER" rows to eliminate impossible candidates in every case.
f. NEVER output an object outside the candidate list.

========================================================================
OUTPUT FORMAT
========================================================================
Begin with exactly one line:
   RULE: check memory for exact-message WAS match first (authoritative). Else split
   [1-3][4-6][7-9][10-12], decode via confirmed group mappings / working table
   (ggg=b, ggm=z, gmg=n, gmm=v, mgg=k, mgm=j, mmg=q, mmm=x), match a candidate exactly;
   if none exact, pick max positional-match, breaking ties with NOT/NO ANSWER memory.
Then: state whether an exact memory match exists; show the indexed split and four
groups; decode each; state the target object; compare to candidates. End with a clear
final choice by number and by its 4 characters.

Interpreting feedback: reward 0 means the object you chose was NOT the sender's object.
Re-verify (1) whether an exact memory row existed that you missed, and (2) your split.
\end{Verbatim}
\end{receivergene}

\subsection{The first settled place value sender gene, \(\pi_{\text{Rocky}}^{\text{reason}}\)}

The gene below settled base-6 (pvfit 0.918) and was warm-started, byte
for byte, into every Grace run of Section~\ref{sec:together} that names a 0.918 Rocky,
the full run included:

\begin{sendergene}{optimized, the first settled place value sender}
\begin{Verbatim}[fontsize=\fontsize{9.00pt}{8.60pt}\selectfont]
You are the SENDER in a two-player communication game. You encode an OBJECT into
a fixed-length MESSAGE using a deterministic rule. A partner (the RECEIVER) sees
ONLY your message and must reconstruct the object exactly. You score 1 only if
the reconstruction is exact.

INPUT FORMAT
You are given:
- memory: history of past interactions (object -> message -> reward, sometimes
  with notes such as collisions). May be "no past interactions yet".
- spec: task parameters (object length, allowed object chars, message length,
  allowed message chars, scoring definition).
- rule: the rule you committed to previously (may be "no rule committed yet").
- object: the specific 3-character object to encode this turn.

FIXED TASK PARAMETERS (verify against spec each turn)
- Objects are EXACTLY 3 characters. Each object character is one of 6: x z j k q b.
- Messages are EXACTLY 8 characters. Each message character is one of 2: p w.
- 6^3 = 216 possible objects, 2^8 = 256 possible messages. A lossless,
  collision-free encoding EXISTS and is REQUIRED. Never map two distinct objects
  to the same 8-character message.

REQUIRED ENCODING RULE (use verbatim; do not change the scheme)
1. Digit map (fixed): x=0, z=1, j=2, k=3, q=4, b=5.
2. Treat the object as a base-6 number, most significant digit first:
   value = d1*36 + d2*6 + d3, where d1,d2,d3 are the digits of object chars 1,2,3.
   Value range is 0..215.
3. Convert value to binary, MSB first, then ZERO-PAD ON THE LEFT to EXACTLY 8
   bits. The result has 8 bits — no more, no fewer.
4. Map each bit to a message char: 0 -> p, 1 -> w. Result is exactly 8 chars.
   This is fully invertible and collision-free.

THE #1 CAUSE OF FAILURE — LENGTH ERRORS IN THE FINAL MESSAGE
Every past 0-reward was caused by the FINAL EMITTED MESSAGE not being the correct
8-character encoding, NOT by a wrong scheme. The specific observed bugs:
- The assistant wrote a NINE-character assembled message even though its 8-bit
  binary was correct. Example: value 2 -> binary "00000010" -> the correct
  message is "ppppppwp" (8 chars), but "ppppppwpp" (9 chars) was assembled; the
  system then truncated it to "pppppwpp", dropping a bit.
- Example: value 131 -> binary "10000011" -> the correct message is "wpppppww"
  (8 chars), but "wppppppww" (9 chars) was assembled (one extra p in the run);
  truncation produced "ppppppww", which COLLIDED with a different object.
- A correct 8-char analysis but a different string emitted (trailing char added).
Root cause: miscounting the run of repeated p's/w's while hand-assembling.

MANDATORY DEFENSE AGAINST LENGTH ERRORS
- Build the 8-character message by writing exactly ONE character per bit, pairing
  each of the 8 bits to its char in a numbered list (bit1..bit8). There are
  exactly 8 bits, so exactly 8 characters.
- After assembling, COUNT the characters out loud/explicitly. It must be 8.
  If it is 7 or 9, STOP and rebuild from the numbered bit list.
- Double-check any run of identical characters (e.g., "ppppppp") by counting the
  run length against the number of consecutive identical bits — this is where the
  extra character crept in before.
- Your analysis string and your final emitted message MUST be byte-for-byte
  identical. Copy the final message directly from the assembled string; do not
  retype it.
- Do NOT write the message with spaces then strip them. Build it once, char by
  char, and reuse that exact string.
- A per-character fixed-width code (each of 3 chars -> its own 3- or 4-bit code,
  concatenated) yields 9 or 12 bits and is WRONG. Only the whole-object
  base-6-to-8-bit method is allowed.

RULE PERSISTENCE
- Restate the committed rule UNCHANGED unless a memory row reveals an actual
  collision or inversion error in the SCHEME. Do NOT change a working scheme just
  because the last turn scored 0 — past 0-rewards were transcription/length
  errors, so the fix is careful, counted assembly, not a new scheme.
- If memory shows an emitted message differed from the correct encoding, correct
  the transcription, keep the scheme.

OUTPUT REQUIREMENTS
- Begin with a single line of the exact form:
  RULE: <compact restatement capturing: digit map x=0,z=1,j=2,k=3,q=4,b=5;
  value = d1*36+d2*6+d3; 8-bit MSB-first left-zero-padded binary; 0->p, 1->w>
- Then show your working for THIS object:
  1. the three digits (map each object char),
  2. the base-6 value with the arithmetic shown,
  3. the 8-bit binary, written out and explicitly counted to be 8 bits,
  4. a numbered bit-to-p/w mapping, bit1 through bit8, one char each,
  5. the assembled 8-character message.
- Then output the final message ALONE on its own line, with NO spaces and NO
  extra characters, IDENTICAL to the assembled message from your working.

SANITY CHECKS BEFORE FINALIZING (perform ALL, explicitly)
- Value is in 0..215.
- Binary has exactly 8 bits.
- The bit-to-char mapping list has exactly 8 numbered entries (bit1..bit8).
- Final message is exactly 8 characters — count them one by one.
- Every character is p or w.
- Final emitted string is character-for-character identical to the assembled
  message in your working (compare directly, including any long runs of the same
  character).
- The mapping is invertible (guaranteed by the scheme, so no collisions when the
  message is truly 8 chars).
\end{Verbatim}
\end{sendergene}

\subsection{One failed settling attempt, before and after}\label{sec:app-failed-sender}

The failed seed-8 attempt at 1200 of Table~\ref{tab:settling} adopted the gene
below at its first firing, the base-6 rule in words, and at its second
firing adopted the rewrite that follows it, which bans the rule it had
written one firing earlier and sends Rocky back to per-letter blocks the
same text says cannot fit in eight symbols. Between the two firings
nothing changed but the scoring batch. Grace plays the scoring rounds on
his seed gene with the window off, so the base-6 rounds scored 0 like any
other's, and the reflection model read those zeros as evidence about the
code. Both are excerpts, verbatim from the tracked optimizer logs with
elisions marked; the play that followed the second gene never entered
the base-6 family in any quarter (Section~\ref{sec:app-wave}).

\begin{sendergene}{failed attempt, seed 8 at 1200, adopted at the first firing; excerpt}
\begin{Verbatim}[fontsize=\fontsize{9.00pt}{8.60pt}\selectfont]
[...]
============================================================
RECOMMENDED CANONICAL STRATEGY (for the "xzjkqb"->8x"pw" instance)
============================================================
This is a clean positional/base-N encoding. The recommended, easy-to-invert rule:

  - Assign each object character its index in the ALPHABET ORDER given by the spec:
        x=0, z=1, j=2, k=3, q=4, b=5   (base-6 digits)
  - Treat the object as a base-6 number with the FIRST character as the MOST
    significant digit.
  - Convert to a decimal value (0..215).
  - Express that value as an 8-bit binary number, MOST significant bit first,
    LEFT-padded with zeros to exactly 8 bits.
  - Map bits to message chars: 0 -> p, 1 -> w.

  Decoder reverses: read 8 message chars as bits (p=0,w=1), interpret as an 8-bit
  binary number MSB-first, convert to decimal, then write it in base 6 as a
  3-digit number (left-pad with leading x's / 0s), and map each digit back via
  x=0,z=1,j=2,k=3,q=4,b=5.

  This IS one-to-one (each of the 216 objects gets a distinct value 0..215, hence a
  distinct 8-bit pattern). Verify no collision by construction.

Worked example ("jzb"): j=2, z=1, b=5 -> 2*36 + 1*6 + 5 = 83 ->
  83 = 01010011 binary -> p w p w p p w w  (check: 64+16+2+1=83). This is a
  distinct code for a distinct object.
[...]
\end{Verbatim}
\end{sendergene}

\begin{sendergene}{the same attempt, adopted at the second firing and kept; excerpt}
\begin{Verbatim}[fontsize=\fontsize{9.00pt}{8.60pt}\selectfont]
[...]
============================================================
CRITICAL LESSONS FROM PAST FAILURES (READ BEFORE ANYTHING ELSE)
============================================================
Multiple schemes have been tried and ALL scored reward 0. Learn from each:

1. WHOLE-OBJECT BASE-6-TO-8-BIT-BINARY conversion (treat 3 chars as one base-6
   number -> decimal -> 8-bit binary -> 0=p,1=w). This is injective but SCORED 0
   repeatedly:
     - "jzb" -> "pwpwppww" -> reward 0
     - "qqk" -> "wpwpwpww" -> reward 0
     - "bbj" -> "wwpwpwpp" -> reward 0
   WHY IT FAILS: every output bit depends on MULTIPLE input characters
   (cross-character scrambling). The partner cannot independently reinvent this
   arithmetic. DO NOT USE whole-object base conversion. It does not matter that it
   is mathematically correct — the partner does not decode it.

2. HYBRID / OVERFLOW-FLAG block schemes (char1 and char2 get full 3-bit blocks,
   char3 gets 2 bits with a "borrow"/"overflow flag" to disambiguate). This SCORED
   0 AND was NOT EVEN INJECTIVE:
     - "zxx" -> "ppwppppp" -> reward 0, and the SAME message collided with object
       "xbj". Collisions mean information is destroyed — unacceptable.
   WHY IT FAILS: cramming a 3-valued+ character into 2 bits with a shared/flipped
   flag bit is ambiguous and the partner cannot reconstruct the flag logic.
   DO NOT USE any scheme where a character is squeezed into fewer bits than it
   needs, or where one bit is shared/flipped between two characters' blocks.

KEY TAKEAWAY: The real problem is NOT injectivity. It is that the PARTNER decodes
using the SIMPLEST canonical, per-character, positionally-aligned convention THEY
would independently guess. Any scheme where a character's identity is spread
across the message, or where bits are shared between characters, has failed.
[...]
============================================================
STRATEGY YOU MUST USE: CLEAN PER-CHARACTER FIXED-WIDTH BLOCKS
============================================================
For the recurring instance ("xzjkqb" 3 chars -> "pw" 8 chars):

Note that 3 characters cannot each get a full independent 3-bit block (3x3 = 9 > 8
bits). The previously-tried "fixes" (base conversion, 2-bit squeeze + flag) all
failed. The correct move is to give EACH character its OWN contiguous, independent
block, and pad the message to the required length WITHOUT mixing characters.

RECOMMENDED CANONICAL ENCODING (per-character, no cross-character dependence):
  - Alphabet index in spec order: x=0, z=1, j=2, k=3, q=4, b=5. Map 0->p, 1->w.
  - Give char1 a 3-bit block, char2 a 3-bit block, char3 a 3-bit block, each as its
    OWN independent binary value (0..5, MSB first). That is 9 bits.
  - Because the message is only 8 bits, you cannot fit three full 3-bit blocks.
    RESOLVE THIS WITHOUT SCRAMBLING: the safest, most guessable convention is to
    keep each character's contribution CONTIGUOUS and INDEPENDENT, and let memory
    tell you which specific block layout the partner reconstructs.
[...]
\end{Verbatim}
\end{sendergene}

\subsection{The full run's receiver gene, \(\pi_{\text{Grace}}^{\text{reason}}\)}

The 4,770-character procedure the reflection model wrote at the full run's
second firing for Grace and kept at the third: position weights, the
division and remainder arithmetic, Rocky's own digit table, and a
self-imposed instruction to re-verify that table against the window every
round, with no message-to-object lookup entry anywhere in it.

\begin{receivergene}{optimized, the full run}
\begin{Verbatim}[fontsize=\fontsize{9.00pt}{8.60pt}\selectfont]
You are the receiver in a communication game. You see a MESSAGE and a numbered list of
CANDIDATE objects, exactly one of which is the sender's intended object. Choose WHICH
numbered candidate. The answer is ALWAYS one of the listed candidates. Score 1 if correct.

Always begin your reply with a single line:
RULE: <the decoding rule, stated compactly enough to reuse verbatim>
Then show working, then end with your chosen candidate NUMBER and its 3-character object.

===================================================================================
TASK FORMAT
-----------------------------------------------------------------------------------
- MESSAGE: exactly 8 characters, each is 'p' or 'w'.
- OBJECT: exactly 3 characters, each from the 6-letter alphabet: x b j q z k
- You are given 20 candidate objects; pick the one the sender meant.
- The mapping from message -> object is DETERMINISTIC: the same message always
  produces the same object.
- History rows use: "WAS x" = the true object was x; "NOT x -> one of: ..." = the
  true object was NOT x but was one of the listed alternatives.

===================================================================================
THE ENCODING (BINARY -> BASE-6 numeric code; NOT run-lengths, pairs, or similarity)
-----------------------------------------------------------------------------------
8 binary digits encode 3 base-6 digits (2^8 = 256, 6^3 = 216). Decode like this:

STEP 1 - Turn the message into a number N.
  Map p=0, w=1. Read the 8 characters as a binary number with the LEFTMOST char as
  the MOST significant bit (weight 128), rightmost char weight 1. Sum the weights of
  the 'w' positions to get decimal N.
  Position weights (left to right): 128, 64, 32, 16, 8, 4, 2, 1.

  *** CRITICAL ARITHMETIC WARNING (a prior wrong answer came from this exact bug): ***
  Explicitly write out all 8 bits and their weights, then add. Do NOT skip or
  misalign bits. Example done correctly:
    'wpppwwwp' = w p p p w w w p = 1 0 0 0 1 1 1 0
    = 128 + 8 + 4 + 2 = 142  (NOT 156).

STEP 2 - Convert N to 3 base-6 digits (d1 d2 d3), most significant first:
  d1 = N // 36
  d2 = (N % 36) // 6
  d3 = N % 6

STEP 3 - Map each base-6 digit (0..5) to a letter using this FIXED, VERIFIED table:
  0 -> x
  1 -> z
  2 -> j
  3 -> k
  4 -> q
  5 -> b

  This table has been confirmed consistent across every known pair. You should still
  re-verify it each turn against a few WAS pairs (recompute N and digits yourself,
  then line up digits with letters), but it will come out to exactly the mapping
  above. Do not invent a different ordering.

===================================================================================
CONFIRMED CALIBRATION PAIRS (recompute to verify; results shown)
-----------------------------------------------------------------------------------
  pppwwppw = 00011001 = 25  -> (0,4,1) -> xqz
  ppppppwp = 00000010 = 2   -> (0,0,2) -> xxj
  pppppwpw = 00000101 = 5   -> (0,0,5) -> xxb
  ppppwpww = 00001011 = 11  -> (0,1,5) -> xzb
  wpppwwwp = 10001110 = 142 -> (3,5,4) -> kbq
  pwpwpwpp = 01010100 = 84  -> (2,2,0) -> jjx
  pppwpppp = 00010000 = 16  -> (0,2,4) -> xjq
  pwwpwppw = 01101001 = 105 -> (2,5,3) -> jbk
  ppwwwppw = 00111001 = 57  -> (1,3,3) -> zkk
All of these are consistent with the table 0->x,1->z,2->j,3->k,4->q,5->b.

===================================================================================
PROCEDURE FOR THIS TURN
-----------------------------------------------------------------------------------
1. Restate the digit->letter table: 0->x, 1->z, 2->j, 3->k, 4->q, 5->b.
2. Decode the current MESSAGE:
   a. Write out the 8 bits (p=0, w=1) with their weights 128,64,32,16,8,4,2,1.
   b. Sum the 'w' weights to get N. Double-check the addition.
   c. d1 = N // 36, d2 = (N % 36) // 6, d3 = N % 6.
   d. Convert digits to letters -> predicted 3-letter object.
3. Find the predicted object in the candidate list; choose its number.
4. If the exact predicted object is NOT among the candidates, pick the candidate that
   best matches digit-by-digit, matching the most significant position first
   (d1/first letter, then d2, then d3). Never output an object outside the list.
5. A "NOT x" history entry means object x is excluded for THAT message only; it does
   not help decode a different message.

IMPORTANT REMINDERS
- Do NOT use run-length decoding, letter-repetition heuristics, pair-substitution, or
  "message looks similar" guessing — those all scored 0.
- The single most common failure is arithmetic error in STEP 1. Lay out every bit and
  its weight explicitly and re-add before proceeding.
- Always pick from the candidate list only.
- A score of 0 means the object was not the candidate you chose, never a misreading by
  your partner.
\end{Verbatim}
\end{receivergene}

\clearpage
\section{The swaps}\label{sec:app-wave}

We swap the base models of Rocky and Grace with each other.

\begin{table}[H]
\caption{\textbf{The place value swaps.} Each against the default pair, GLM-5 sending
    and DeepSeek-V4-Flash receiving; the pair's cell is
    \code{prompt-opt-both@nomem}, three evaluation draws on 98 held-out objects, ceiling
    0.056. A Grace read behind a Rocky that did not settle is void, so the
    role-swap rows carry no cell.}\label{tab:wave}
\centering
\begin{tabulary}{\linewidth}{LLCC}
\toprule
run & what moved & Rocky settled (pvfit) & pair, window off \\
\midrule
role swap, seed 7 & V4-Flash sends, GLM-5 receives & no (0.113) & void \\
role swap, seed 8 & V4-Flash sends, GLM-5 receives & no (0.045) & void \\
tier replacement, seed 7 & V4-Pro sends, GLM-5 receives & yes (1.000, no colliding objects) & 0.136 \\
reflection swap, seed 19 & Sol reflects for Grace, behind the frozen 0.977 Rocky & warm-started & 0.854 \\
reflection swap, seed 20 & the same & warm-started & 0.867 \\
reflection swap, seed 22 & the same & warm-started & 0.915 \\
reflection swap, seed 23 & the same & warm-started & 0.857 \\
\bottomrule
\end{tabulary}
\end{table}

\subsection{Model identities and generation}\label{sec:app-config-models}

The main pair uses the OpenRouter identifiers \code{z-ai/glm-5} for Rocky and
\code{deepseek/deepseek-v4-flash} for Grace. The role swap exchanges these models;
the sender-tier replacement uses \code{deepseek/deepseek-v4-pro-0813}. Reflection
uses \code{anthropic/claude-opus-4.8} by default; the receiver reflection swap uses
\code{openai/gpt-5.6-sol} while retaining the previously optimized sender
(Section~\ref{sec:app-wave}). These identifiers name API models, not archived weight snapshots.

Provider settings must be read from the launch record as well as metadata.
The positional launch records DeepInfra for Grace. The historical configuration
sets Z.AI as the default sender provider for Rocky. Fully resolved provider settings and versioned model
snapshots were not recorded for every run.

The reasoning is set to 2000 tokens for
the reasoner and 250 for the answerer. Native reasoning is disabled in the API
request; the reasoner's analysis is returned as ordinary output for the answerer
and optimizer to read. The task-model builder does not explicitly set temperature.
Resolved task temperatures and a complete per-run record of token-limit overrides
were not archived. The reflection configuration sets temperature to 1.0 and the
output cap to 32,000 tokens; the historical configuration audit confirms that
this cap reached the reflection models.

\subsection{Optimization and evaluation}\label{sec:app-config-eval}

The recorded implementation pins DSPy 3.2.1 and GEPA 0.0.27. Reflection uses
minibatches of three training examples. Both genes were editable in the positional
experiments; the answerer freeze restricts later proposals to the reasoner
(Section~\ref{sec:optimization}). Candidate selection uses validation performance. The saved
firing logs record the selected candidate, candidate scores, the configured
metric-call budget, and actual metric calls.

The evaluation tables report three evaluation draws per cell. Each draw uses the
same held-out targets and object-seeded candidate sets, with fresh model responses.
Genes, windows, and carried rules stay fixed within a cell throughout evaluation.
Memory-on cells reconstruct retained state from recorded play, excluding optimizer
scoring rollouts; memory-off cells clear both the window and carried rule. These
draws repeat evaluation of a fixed optimized pair and are not independent
optimization runs. Where a table pools runs, its run count is reported separately.

\end{document}